\documentclass[10pt,twocolumn,letterpaper]{article}

\usepackage{times}
\usepackage{epsfig}
\usepackage{graphicx}
\usepackage{subcaption}
\usepackage{amsmath}
\usepackage{amssymb}
\usepackage{arydshln}
\usepackage[accsupp]{axessibility}

\newcommand{\argmax}{\mathop{\mathrm{argmax}}}
\usepackage[pagenumbers]{cvpr} 

\usepackage[dvipsnames]{xcolor}
\newcommand{\camerareadychange}[1]{{\color{black}#1}}

\definecolor{cvprblue}{rgb}{0.21,0.49,0.74}
\usepackage[pagebackref,breaklinks,colorlinks,citecolor=cvprblue]{hyperref}
\def\paperID{15322} 
\def\confName{CVPR}
\def\confYear{2024}

\newcommand{\setNotation}[1]{\mathcal{#1}}
\newcommand{\setIm}{\setNotation{I}}
\newcommand{\setRef}{\setNotation{R}}
\newcommand{\setPose}{\setNotation{P}}
\newcommand{\setMap}{\setNotation{M}}
\newcommand{\kNN}{K} 
\newcommand{\setRefNN}{\setRef_{\textrm{nn}}}
\newcommand{\setPoseNN}{\setPose_{\textrm{nn}}}
\newcommand{\setImTrain}{\setIm_{\textrm{train}}}

\newcommand{\setPoseTrain}{\setPose_{\textrm{train}}}

\newcommand{\Gauss}{\mathbf{N}} 
\newcommand{\scale}{\lambda} 

\title{On the Estimation of Image-matching Uncertainty in Visual Place Recognition}

\author{Mubariz Zaffar\\
ME, TU Delft\\
The Netherlands\\
{\tt\small M.Zaffar@tudelft.nl}
\and
Liangliang Nan\\
ABE, TU Delft\\
The Netherlands\\
{\tt\small Liangliang.Nan@tudelft.nl}
\and
Julian F. P. Kooij\\
ME, TU Delft\\
The Netherlands\\
{\tt\small J.F.P.Kooij@tudelft.nl}}

\begin{document}

\maketitle

\begin{abstract}
In Visual Place Recognition (VPR) the pose of a query image is estimated by comparing the image to a map of reference images with known reference poses. As is typical for image retrieval problems, a feature extractor maps the query and reference images to a feature space, where a nearest neighbor search is then performed. 
However, till recently little attention has been given to quantifying the confidence that a retrieved reference image is a correct match. Highly certain but incorrect retrieval can lead to catastrophic failure of VPR-based localization pipelines. This work compares for the first time the main approaches for estimating the image-matching uncertainty, including the traditional retrieval-based uncertainty estimation, more recent data-driven aleatoric uncertainty estimation, and the compute-intensive geometric verification. We further formulate a simple baseline method, ``SUE'', which unlike the other methods considers the freely-available poses of the reference images in the map. Our experiments reveal that a simple L2-distance between the query and reference descriptors is already a better estimate of image-matching uncertainty than current data-driven approaches. SUE outperforms the other efficient uncertainty estimation methods, and its uncertainty estimates complement the computationally expensive geometric verification approach. Future works for uncertainty estimation in VPR should consider the baselines discussed in this work. 
\end{abstract}

\section{Introduction}

\begin{figure}
\begin{center}
\includegraphics[width=0.9\linewidth]{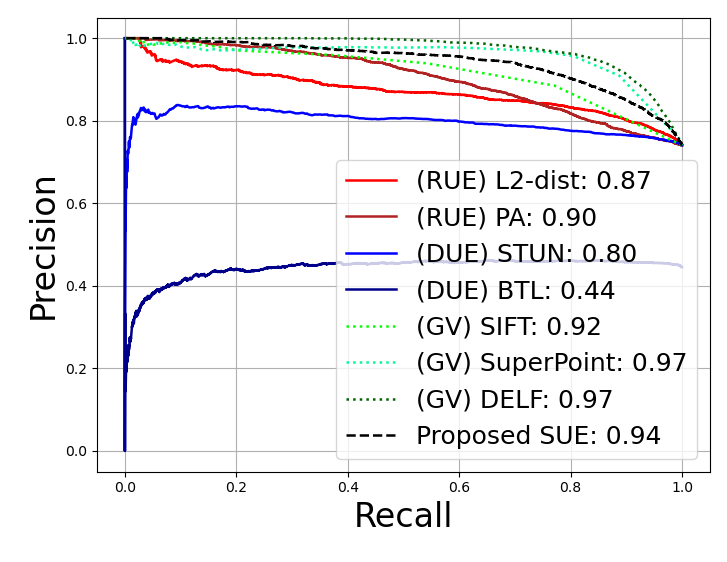}
\end{center}
   \caption{
The Precision-Recall curves on the Pittsburgh dataset~\cite{arandjelovic2016netvlad} for the three common categories of VPR uncertainty estimation methods (\color{red}{RUE}, \color{blue}{DUE}, \color{green}{GV}\color{black}), 
and for our proposed baseline \textbf{SUE} which uniquely considers spatial locations of the top-K references.
The global image descriptors~\cite{cai2022stun} are fixed
for all methods except BTL~\cite{warburg2021bayesian}.
\camerareadychange{The \textit{only} difference is the confidence given by each uncertainty estimation method to the best-matched reference descriptors for the corresponding queries.}
The legend lists the Area-under-the-Precision-Recall-curves.
As \textit{GV} methods are \textit{two to three orders of magnitude} more computationally expensive than the others, they are plotted as dotted lines.
Surprisingly, simple L2-distance in feature space is a better estimate of VPR uncertainty than recent deep learning-based uncertainty estimates. SUE outperforms all other efficient uncertainty estimation methods. }
\label{fig:prcurvespitts}
\end{figure}

Visual Place Recognition (VPR) is the problem of identifying a previously visited place given a query camera image and a map of geo-tagged reference images~\cite{lowry2015visual}.
It has applications in vehicle localization~\cite{zhu2018visual}, 3D modeling~\cite{agarwal2011building}, image search~\cite{tolias2016image}, and loop-closure in Simultaneous Localization and Mapping (SLAM)~\cite{lowry2015visual,cadena2016past}.

VPR is typically approached as an image retrieval problem, 
transforming images into feature vectors in a latent feature space where an efficient nearest neighbor search compares the query to all references.
The pose of the query image is then approximated to be the same as that of the retrieved nearest neighbor references.
%
Since successful VPR requires a good image representation that is robust to viewpoint and/or appearance changes~\cite{garg2021your, masone2021survey, lowry2015visual},
the field has benefited from advances in deep representation learning.

However, two images with similar visual content could still originate from geographically far-apart areas, a concept referred to as \textit{perceptual-aliasing} in VPR~\cite{garg2021your}.
For example, images with mostly sky could match many locations on an outdoor map.
This constitutes aleatoric uncertainty, i.e.,~inherent noise or ambiguity in the data which cannot be reduced,
as opposed to epistemic uncertainty which could be addressed with more training data~\cite{kendall2017uncertainties}.
The close proximity of perceptually aliased images in the feature space can result in catastrophic failures
For instance, a highly confident false-positive from VPR could result in an incorrect loop closure in a SLAM pipeline, leading to misaligned maps~\cite{lowry2015visual,cadena2016past}.
Reliable uncertainty estimation on the quality of the match is therefore key to avoid such failures by, e.g.,~rejecting results above a certain uncertainty threshold.
Moreover, uncertainty estimation can also be used to fuse multiple predictions in VPR ensemble methods~\cite{cai2022stun}.


From existing literature, we identify three categories of methods to estimate image-matching uncertainty in VPR: retrieval-based uncertainty estimation (\textit{RUE}), data-driven aleatoric uncertainty estimation (\textit{DUE}), and geometric verification (\textit{GV}) by local feature matching. \textbf{RUE:} \camerareadychange{Traditionally in VPR, the L2-distance between the query and the best-matched reference in the feature space has been used as an estimate of uncertainty~\cite{piasco2018survey}.} The ratio of L2-distance between the first and second nearest neighbour reference is also used~\cite{hausler2021unsupervised}. 
\textbf{DUE:} On the other hand, several recent works, such as the Bayesian Triplet Loss~\cite{warburg2021bayesian} and the Self-teaching Uncertainty Estimation~\cite{cai2022stun}, have proposed to explicitly learn to predict the aleatoric uncertainty from the query's image content only.
\textbf{GV:} Another way to assert matching confidence is to test for consistent geometry among matched local features between the query and the best matching reference in a RANSAC loop~\cite{noh2017large}.

Remarkably,
none of the three categories exploit the spatial locations of matched images in the actual reference map,
which we hypothesize can be an important source of information for estimating VPR matching uncertainty.
To test this hypothesis, we formulate a new simple baseline,  Spatial Uncertainty Estimation (SUE). SUE is a straightforward and efficient approach to estimating uncertainty for a query image's match,
using the spatial spread of the physical poses for the most similar references in the map as a proxy. 
A high spatial spread indicates perceptual aliasing leading to high matching uncertainty, while a low spread indicates a distinct area is matched. An overview of the sources of information employed by all categories of methods and by SUE is provided in Table~\ref{tab:resources_overview_all}.

While all categories of uncertainty estimation methods aim for the same task, i.e., rejecting false positives in VPR, 
previous evaluations did not include all categories, 
providing an incomplete picture of the state-of-the-art.
This work therefore compares the three existing categories and SUE on a levelled playing field, to provide recommendations for future research, and insights on the strengths/weaknesses of each category.
For instance, as the preview of the experimental results in Fig.~\ref{fig:prcurvespitts} indicates,
SUE outperforms other efficient methods (this and other experiments will be discussed in more detail in Section~\ref{sec:experiments}).


\begin{table}
    \centering
    \begin{tabular}{|c|c|c|c|c|}
         \hline
         \textbf{Categ.}  & \textbf{Descr.?} & \textbf{Poses?} & \textbf{Images?} & \textbf{Efficient?}\\
         \hline
         \hline
         RUE & Top-K &  No & No & Yes\\
         \hline
         DUE & No &Only train& Yes & Yes\\
         \hline
         GV & No & No & Yes & No\\
         \hline
 
         \textit{SUE} & \textit{Top-K} & \textit{Top-K} & \textit{No} & \textit{Yes}\\
         \hline
    \end{tabular}
    \caption{Overview of the sources of information needed by the current main categories for VPR uncertainty estimation, and by the proposed method \textit{SUE}:
    the query/reference global image descriptors, the reference poses, or complete query/reference images. 
 Efficiency refers to the inference time needed by each approach.}
    \label{tab:resources_overview_all}
\end{table}

Concretely, our contributions are:

\begin{enumerate}
    \item A comparison of three different categories of uncertainty estimation methods in VPR.

    \item A new simple baseline method, SUE, that considers the spatial locations of the reference images, a source of information not used by existing categories.

    \item Since \textit{GV} gives the best uncertainty estimates albeit at a higher computational cost, we investigate whether the other methods are complementary to \textit{GV}. 
\end{enumerate}

\section{Related work}
\camerareadychange{Visual place recognition was first surveyed in the seminal work of Lowry \textit{et al.}~\cite{lowry2015visual}. The three fundamental VPR challenges identified by Lowry \textit{et al.} are viewpoint changes, appearance changes, and perceptual-aliasing.}

The concept of matching images for VPR dates back to before the deep-learning era, when handcrafted methods were used to perform VPR~\cite{se2002mobile, stumm2013probabilistic, ho2007detecting, cummins2008fab}. However, with the rise of deep learning, many deep learning-based methods were proposed to solve the first two challenges in VPR. A broad categorization of these methods can be done based on their underlying novelty, such as the use of a novel loss function~\cite{revaud2019learning, leyva2021generalized, thoma2020geometrically}, better training data~\cite{berton2022rethinking, ali2022gsv}, new architectures~\cite{wang2022transvpr, zhang2021vector, yu2019spatial}, and new methods for feature aggregation~\cite{arandjelovic2016netvlad, radenovic2018fine, hausler2021patch,ali2023mixvpr}.
A number of benchmarks have been proposed in VPR, for example, the recent Deep Visual Geo-localization benchmark~\cite{berton2022deep}, VPR-Bench~\cite{zaffar2021vpr} and similar benchmarks in the image retrieval community~\cite{Radenovic_2018_CVPR, sattler2018benchmarking}. From these benchmarks, it is clear that the deep learning-based VPR techniques outperform handcrafted techniques by a significant margin on most datasets. 

We focus on the third challenge identified in Lowry~\textit{et al.}, i.e., perceptual aliasing, which arises from aleatoric uncertainty in the data. This challenge has received less attention in VPR literature compared to viewpoint and appearance changes. Most works in VPR use the distance (e.g., L2 or Cosine) in feature space between a query and the nearest neighbor as the uncertainty estimate~\cite{berton2022deep}, or the distance between the retrieved nearest neighbors~\cite{hausler2021unsupervised}. Some more recent works model the aleatoric uncertainty in image retrieval, e.g., the Bayesian Triplet Loss (BTL)~\cite{warburg2021bayesian} and the Self-Teaching Uncertainty Estimation (STUN)~\cite{cai2022stun}. Both BTL and STUN estimate the aleatoric uncertainty in the training data by representing images as distributions instead of point estimates in the feature space. 
Each image thus has an associated mean and variance for a feature descriptor. 

Gronat \textit{et al.}~\cite{gronat2013learning} treat VPR as a classification problem by training place-specific classifiers, one for each place, where each classifier naturally outputs a confidence estimate for the corresponding pose.  Pion~\textit{et al.}~\cite{pion2020benchmarking} approximate the pose of the query image by aggregating the pose hypotheses from the top-retrieved nearest neighbors, weighing each hypothesis based on the distance in the feature space. The variance of the aggregated pose represents uncertainty over the pose space. Notably, this concept of pose uncertainty has been modeled in these existing works~\cite{pion2020benchmarking, gronat2013learning, zaffar2023copr} and other related tasks such as classical Particle Filters~\cite{dellaert1999using}, but, to the authors' best knowledge, the uncertainty estimates derived based on the distribution of pose hypotheses has not yet been studied as a proxy for image-matching uncertainty.

Beyond global descriptors-based VPR, in local feature matching-based image retrieval the inlier count (aka. geometric verification) has been used as an estimate of confidence~\cite{noh2017large}.
Zeisl~\textit{et al.}~\cite{zeisl2015camera} perform 2D-to-3D local feature matching to estimate a distribution over the possible query poses. The work of \cite{zamir2010accurate} uses such inlier count from local feature matching and combines it with the pose distribution of retrieved images to estimate the confidence of localization. \camerareadychange{Since local features can appear in similar geometric configurations (geometric burstiness) across unrelated images, \cite{sattler2016large} proposes to use the pose information to downweight such matches in the inlier count. However, retrieving images based on local feature descriptors is computationally expensive, whereas VPR instead only efficiently compares global image descriptors. \footnote{We study the relation between geometric burstiness and SUE in the supplementary materials. }} \camerareadychange{Absolute Pose Regression (APR) directly regresses the absolute pose given a camera image, and has also considered pose uncertainty estimation. Some approaches to uncertainty-aware APR include CoordiNet~\cite{moreau2022coordinet}, Bayesian PoseNet~\cite{kendall2016modelling}, and HydraNet~\cite{peretroukhin2019probabilistic}.
Unlike VPR, APR approaches do not generalize to new environments. In this work, we focus on estimating the image-matching uncertainty for VPR. }

\section{Methodology}
This section first introduces the task of uncertainty estimation for VPR.
We then formalize VPR, and describe the three main categories of uncertainty estimation methods. Next, we formulate the proposed baseline approach, SUE, which unlike the other three categories uses the freely available reference poses information. Finally, we outline how we combine the different categories of methods with the computationally expensive geometric verification to investigate if the uncertainty estimates are complementary.

\subsection{Uncertainty estimation in VPR}
\label{sec:AleatoricUncertaintyinVPR}

Typically VPR is considered as an image retrieval task: finding the most similar reference images to the query by Euclidean distance in some feature space.
The poses associated with the images however distinguish VPR from other image retrieval tasks, such as web search, where matches are correct if their image content should be judged as the same.
In VPR we often instead refer to the location of the query and references to judge matches:
a retrieved reference is only acceptable if its pose is within a maximum distance threshold of the (unknown) true pose of the query~\cite{berton2022deep, zaffar2021vpr, garg2021your}.
Ideally, the closest matches in the feature space thus also have the poses closest to the query pose.
However, this is often not the case in VPR due to \textit{perceptual aliasing}, 
a form of aleatoric uncertainty
since it cannot be reduced by choosing a different feature encoder or by using more training data.

It is therefore desirable to obtain some \textit{uncertainty score} $s_q$
for a query and the retrieved nearest neighbor,
where a low score expresses confidence that the nearest neighbor is a correct match.
A threshold $\tau$ on the score could then reject a query ($s_q > \tau$) for which the best match is at risk of being incorrect
to prevent failures 
of the downstream application~\cite{garg2021your}.
The objective of VPR uncertainty estimation is thus to score queries, such that queries with reliable matches 
can be distinguished from those with possible incorrect matches.
Note that while an uncertainty estimation method could provide scores with an explicit probabilistic interpretation,
this is not a strict requirement to apply an acceptance threshold.

\subsection{Formalizing VPR}
Given a set of reference images $\setIm$ with known poses $\setPose$, 
the goal of VPR is to find one or multiple reference images $I_i \in \setIm$ that match the place of a query image $I_q$.
The unknown pose $p_q$ for the query $I_q$ can then be approximated from the poses of the matched references $p_i \in \setPose$, since correct matches should have been taken in the same area.
The exact formulation of a pose generally depends on the localization source and the task, for example, 2D GPS coordinates for visual geo-localization~\cite{berton2022deep}, or 6D pose~\cite{laskar2017camera}.
In this research, we follow a general task-independent formulation and only assume that a pose $p_i$ consists of 2D or 3D spatial coordinates in some global coordinate system.

In the offline map preparation phase of VPR, before accepting queries, 
a feature extractor $G$ is applied to every reference image $I_i \in \setIm$
to obtain $D$-dimensional reference feature descriptors $f_i = G(I_i)$.
Usually $G$ is a trained neural network~\cite{masone2021survey} or a handcrafted feature descriptor~\cite{dalal2005histograms}.
The resulting VPR map $\setMap = (\setRef,\setPose)$
contains the reference feature descriptors set $\setRef = \{ f_1, \cdots f_N \}$, where each descriptor $f_i$ is associated with a corresponding pose $p_i \in \setPose$.

At test time, the same feature extractor $G$ is applied to the query image $I_q$, and its query descriptor $f_q = G(I_q)$ is compared to the reference descriptors in the map $\setMap$.
This can be achieved through an efficient $K$-nearest neighbor lookup,
considering the L2-distances $d_i = || f_{i} - f_q ||_2$ between each reference $i$ and the query.
This gives an ordered list of $\kNN$ nearest neighbor references $\setRefNN = [f_{(1)}, \cdots, f_{(\kNN)}]$,
ranked by increasing distance $d_{(1)} \le \cdots \le d_{(\kNN)}$
and with corresponding poses $\setPoseNN = [p_{(1)}, \cdots, p_{(\kNN)}]$.
Here we use bracketed subscript $(j)$ to indicate $j$-th item in the ranked order, i.e.,~$f_{(1)} = \textrm{argmin}_{f_i \in \setRef} || f_{i} - f_q ||_2$ is the descriptor with the smallest distance to the query in the feature space.

Each corresponding pose $p_{(i)} \in \setPose$ can be considered as a hypothesis to estimate the query's true pose $p_q$, though usually only the pose of the best matching reference feature descriptor $f_{(1)}$ is considered as the VPR pose estimate $p'_q$ for the query, i.e., $p'_q=p_{(1)}$~\cite{zaffar2021vpr}. 
We follow this best-match-based query pose estimation in this work.
In benchmarks, a match is considered correct if $p'_q$ is `physically near' to $p_q$.
The threshold on what distance is still accepted as the same `place' depends on the scale of each localization task~\cite{lowry2015visual}.

\subsection{Current VPR uncertainty estimation categories}
\label{sec:currentvpruncertaintyestimationmethods}

We now describe various representative uncertainty estimation methods for the three common categories.

\textbf{Retrieval-based uncertainty estimation (\textit{RUE}):}
Commonly, the matching uncertainty in VPR is considered proportional to the L2-distance from the best match $d_{(1)}$, so $s_q = d_{(1)}$~\cite{berton2022deep, zaffar2021vpr}, as this distance indicates relevant differences between the visual content of the query and match.

An alternative is to consider the distance ratio between the first and second nearest neighbor, $s_q = d_{(1)}/d_{(2)}$.
This ratio is quite similar to the perceptual aliasing score (PA score)~\cite{hausler2021unsupervised} and the false-positive rejection criterion in the popular local feature descriptor SIFT~\cite{lowe2004distinctive}.

\textbf{Data-driven uncertainty estimation (\textit{DUE}):}
State-of-the-art VPR encoders are typically deep neural networks trained on a labeled VPR dataset. The labeled training data contains the ground-truth poses $\setPoseTrain$ for the training references and query images $\setImTrain$.
A deep encoder $G$ can be adapted to also predict the aleatoric uncertainty of matching a nearby pose,
by learning from the training query image in $\setImTrain$ when an image is distinctive and obtains good pose matches within $\setPoseTrain$, and when not (e.g., images of trees, uniform walls, or sky).
Methods in this category include the Bayesian Triplet Loss (BTL)~\cite{warburg2021bayesian},
and STUN~\cite{cai2022stun}.
Note that the learned uncertainty is based on the training images and poses, not those in the test-time reference map $\setMap$.

In general, an uncertainty-aware encoder $(\bar{f}_i, \sigma^2_i) =  G'(I_i)$
predicts for an image $I_i$ not only the expected feature $\bar{f}_i$, but also the \textit{variance in the feature space}, i.e.,~$f_i \sim \Gauss(\bar{f}_i, \sigma^2_i)$.
The total variance in $\sigma^2_i$ can be used as a proxy for the image-matching uncertainty, $s_q = ||\sigma^2_i||_1$. 
The computational overhead of the deep network producing an additional output $\sigma^2_i$ is low.

\begin{figure}
\begin{center}
\includegraphics[width=1.0\linewidth]{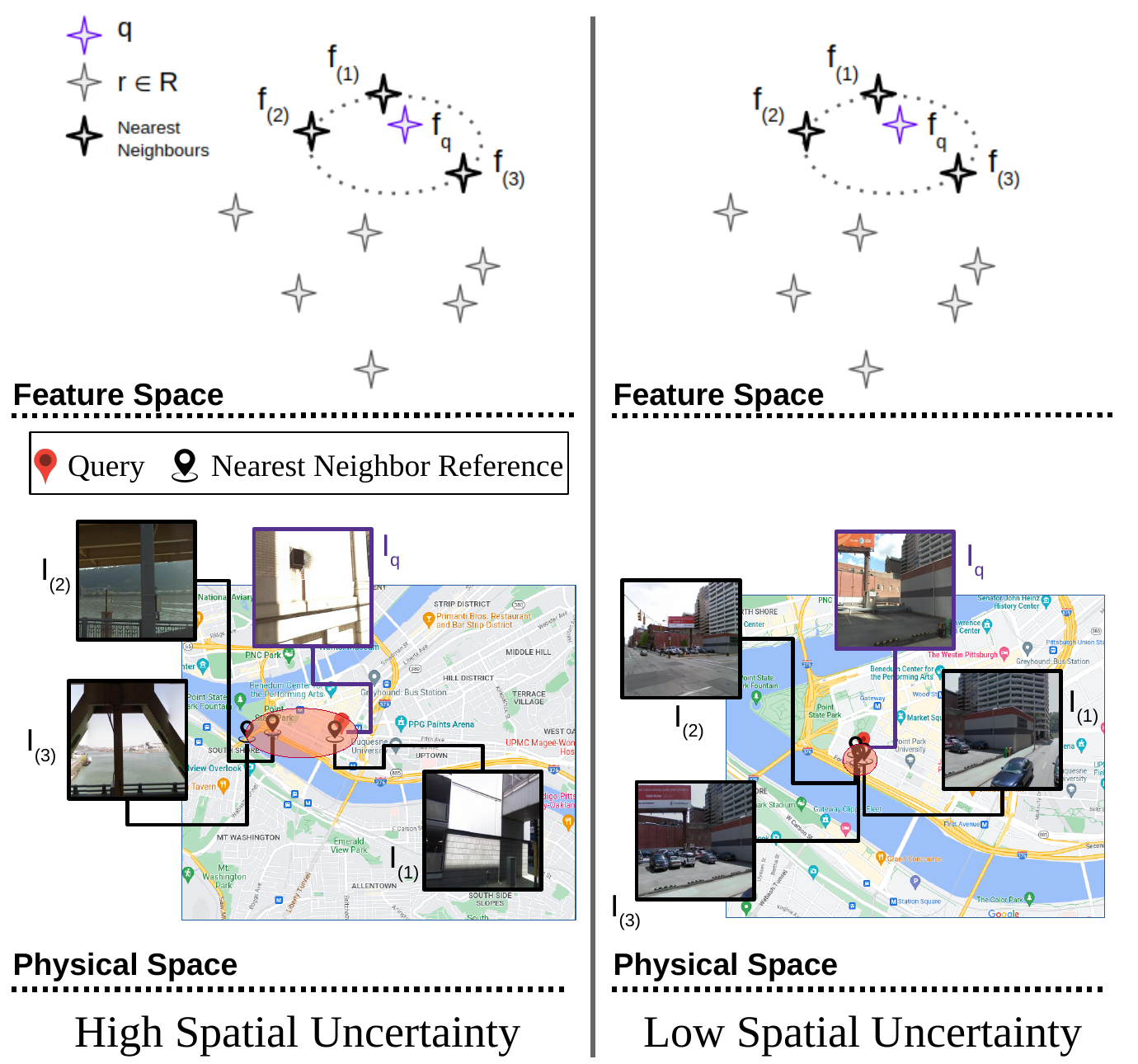}
\end{center}
   \caption{
   In VPR, a query $q$ is compared in feature space to features $f_i \in \setRef{}$ of reference images with known poses.
   The nearest neighbors $f_{(1)}, \cdots, f_{(\kNN)}$ are retrieved as matches.
   Left: The retrieved references $I_{(1)}, I_{(2)}, I_{(3)}$
   share similar visual content with the query (walls, pillars, and blobs),
   but are geographically far apart,
   reflecting high uncertainty that the matched reference is correct.
   Right: For another query, the retrieved references are geographically close together, indicating low uncertainty. 
   }
\label{fig:uncertainty_in_vpr_fig1_combined}
\end{figure}

\textbf{Geometric verification (\textit{GV}):}
Another way to estimate image-matching uncertainty is to compare the query and the best-matched reference image in more detail through local feature matching and geometric verification in a RANSAC loop, e.g., through the use of SIFT~\cite{lowe2004distinctive}, DELF~\cite{noh2017large}, and SuperPoint~\cite{detone2018superpoint}.
All the matched local features that satisfy a geometric transformation estimated from the randomly sampled set of matched local features between the query image and the reference image are considered inliers.
The confidence is indicated by the number of inliers $c_{gv}$, which could be expressed as a matching uncertainty estimate, i.e., $s_q = -c_{gv}$.
While geometric verification yields high-quality uncertainty estimates, 
such post-processing is computationally expensive
compared to the other methods.

\subsection{Spatial uncertainty estimation (\textbf{\textit{SUE}}) for VPR}
We observe that the poses in the reference set $\setPose$ are a potentially powerful and freely available source of information at \textit{test time}, which current uncertainty estimation methods do not exploit
(more details will be presented in Sec.~\ref{sec:AleatoricUncertaintyinVPR}).
The intuition behind this is illustrated in Fig.~\ref{fig:uncertainty_in_vpr_fig1_combined}, where we show that if the nearest neighbors in the feature space are spatially far apart in their respective 2D/3D world coordinates, 
it indicates that such a feature suffers from perceptual aliasing:
various areas in the test reference set contain the queried appearance, thus uncertainty on the pose estimate should be high.
On the other hand, if the nearest neighbors in the feature space are also spatially close together, there is agreement among the matching pose hypotheses that the matched area is distinct within that given reference set, thus the uncertainty should be low.

To test this insight, we now formulate SUE, a purposefully simple image-matching uncertainty estimation method.
Given the $\kNN$-best retrieved references, fit a 2D or 3D multivariate Gaussian distribution $\Gauss(\mu_{p}, \Sigma_{p})$ over their 2D or 3D poses $\setPoseNN$,
\begin{align}
    \mu_{p} &= 
    \frac{1}{\sum_i w_{(i)}} \sum_{i=1}^{\kNN} w_{(i)}\cdot p_{(i)}, \label{eq:1}
\\
    \Sigma_{p} &=
    \frac{1}{\sum_i w_{(i)}}
    \sum_{i=1}^{\kNN} w_{(i)} \cdot (p_{(i)} - \mu_{p}) (p_{(i)} - \mu_{p})^\top, \label{eq:2}   
\end{align}
where the relative contribution $w_{(i)}$ of the $i$-th best reference pose $p_{(i)}$ decreases as 
its L2-distance $d_{(i)}$ to the query in the feature space increases,
\begin{align}
    w_{(i)} &= {e ^ {-\scale \cdot d_{(i)}}}, 
\quad \textrm{where} \quad
    d_{(i)} = ||f_q - f_{(i)}||_2.
    \label{eq:3} 
\end{align}
The total variance across the spatial pose dimensions could then serve as a proxy for image-matching uncertainty, i.e., $s_q = \textrm{trace}(\Sigma_p)$.

The hyper-parameter $\scale$ controls the non-linear relative contribution of a pose $p_i$ for the nearest neighbor $f_{(i)} \in \setRefNN$ given its distance $d_{(i)}$ in the feature space.
This hyper-parameter can be optimized on training data, though our experiments will show that its choice is remarkably robust across various real-world benchmark datasets.
\subsection{Complementing geometric verification}
\label{sec:combiningUEwithGV}
To study to what extent SUE's (or another method's) $s_q$ provides information not captured by the $c_{gv}$ metric from geometric verification,
we treat both scores as a 2D feature vector and train a classifier to predict if a best-matched reference should be accepted as a true-positive, or rejected as a false-positive.
The regular rejection threshold is extended from a single score ($s_q > \tau$) to
a linear weighted sum of both scores ($s_q/\tau_1 + c_{gv}/\tau_2 > 1$), by the use of a regular linear Support Vector Machine (SVM) as a classifier.

\section{Experiments}
\label{sec:experiments}
We first present the setup for our experiments. Then, we compare the performance of all the image-matching uncertainty estimation methods on multiple benchmark datasets. Next, we test if the methods are complementary to geometric verification. Finally, we present an ablation over the hyper-parameters of SUE and provide a discussion.

\begin{table*}
\begin{center}
\begin{tabular}{|l|c|c|c|c|c|c||c|c|}
\hline
Method & $\uparrow$ Pitts. & $\uparrow$ Sanfr. & $\uparrow$ Stluc. & $\uparrow$ Eyn. & $\uparrow$ MSLS & $\uparrow$ Nordland & $\uparrow$ Average & $\downarrow$ Time\\ 
\hline\hline
\textit{(RUE)} L2-distance &0.87 &0.76 &0.79 &0.87 &0.64 &0.18 &0.69 & \textbf{0.05}\\
\textit{(RUE)} PA-Score~\cite{hausler2021unsupervised} &0.90 &0.65 &0.77 &0.88 &0.68 &0.21 &0.68 & \textbf{0.05}\\
\textit{(DUE)} BTL~\cite{warburg2021bayesian} &0.44 &0.17 &0.34 &0.45 &0.21 &0.07 &0.28 & 0.20\\
\textit{(DUE)} STUN~\cite{cai2022stun} &0.79 &0.57 &0.66 &0.71 &0.44 &0.05 &0.54 & 0.10\\
SUE &\textbf{0.94} &\textbf{0.84} &\textbf{0.88} &\textbf{0.93} &\textbf{0.77} &\textbf{0.26} &\textbf{0.77} & 1.08\\
\hline
\textit{(GV)} SIFT-RANSAC~\cite{lowe2004distinctive} &0.92 &0.89 &0.93 &\textbf{0.96} &0.70 &0.15 &0.76 & \textbf{129}\\
\textit{(GV)} DELF-RANSAC~\cite{noh2017large} &0.97 &\textbf{0.92} &\textbf{0.97} &0.95 &\textbf{0.95} &\textbf{0.84}  &\textbf{0.93} & 1587\\
\textit{(GV)} Super-RANSAC~\cite{detone2018superpoint} &0.95 &\textbf{0.95} &\textbf{0.97} &\textbf{0.96} &0.87 &0.50 &0.87 & 848\\
\hline
\end{tabular}
\end{center}
\caption{\camerareadychange{The AUC-PR of all the compared methods. Higher AUC-PR is better, and best is in Bold.
The bottom rows are the computationally expensive geometric verification methods.
The last column lists the time (msec) to give an uncertainty estimate for a single query image.}}
\label{tab:aucprandtime}
\end{table*}

\subsection{Experimental setup}
This section describes the datasets, baselines, evaluation metrics, and implementation details of our work.

\textbf{Datasets:} We use \camerareadychange{six public VPR datasets in this work: Pittsburgh-250k~\cite{arandjelovic2016netvlad}, Sanfrancisco~\cite{chen2011city, torii2019large}, Stlucia~\cite{milford2008mapping}, Eysham~\cite{cummins2009highly}, MSLS~\cite{warburg2020mapillary} and Nordland~\cite{nordlanddataset}. Details of these datasets and their respective ground-truths in~\cite{berton2021viewpoint}.}

\textbf{Baselines}: Our primary baselines for uncertainty estimation include the L2-distance in feature space $d_q$, the perceptual aliasing score (PA score~\cite{hausler2021unsupervised}), the Bayesian Triplet Loss (BTL)~\cite{warburg2021bayesian} and STUN~\cite{cai2022stun}.
As the code for BTL is not open-source, we implement it following the pseudo-code and the network details provided in the original paper.

For geometric verification, we test three types of local feature descriptors, namely the handcrafted SIFT~\cite{lowe2004distinctive}, the deep-learning-based DELF~\cite{noh2017large}, and SuperPoint~\cite{detone2018superpoint}, which we refer to as SIFT-RANSAC, DELF-RANSAC and Superpoint-RANSAC, respectively.

\textbf{Evaluation metrics:}
The precision-recall (PR) curves have been widely used in VPR for estimating the retrieval quality~\cite{zaffar2021vpr}. However, they can also be used to estimate the uncertainty estimate in VPR as widely used in existing uncertainty estimation tasks in deep learning~\cite{malinin2018predictive, hendrycks2016baseline}.
The choice of PR-curves over the Receiver Operating Characteristic (ROC) curve is due to the absence of true-negatives in employed VPR datasets. Given a fixed list of retrieved images, the Precision-Recall curves can reflect the technique with the better uncertainty estimates $s_q$. A technique that can perfectly classify between true-positives (TP) and false-positives (FP), given the uncertainty estimates, achieves an Area-under-the-Precision-Recall-Curve (AUC-PR) of 1.  

For the combination of uncertainty estimates with geometric verification, the task is formulated as binary classification and we use accuracy as an evaluation metric based on the ground-truth true-positives and false-positives~\cite{berton2021viewpoint}.

\textbf{Implementation details:}
For SUE and all the other baselines except BTL,
we use a ResNet-50 backbone with GeM pooling trained in a self-teaching manner in~\cite{cai2022stun} on the training split of the Pittsburgh dataset.
Each feature vector $f_i$ is 2048 dimensional.
For BTL, the same backbone and training data are used, but using the training procedure specified in the original BTL paper~\cite{warburg2021bayesian}. 
For DELF and SuperPoint, the implementations are open-sourced by the respective authors, and the default settings are employed.
For SIFT-RANSAC we use the OpenCV implementation with the number of extracted features set to 5000, the Lowe test ratio to 0.6, and the number of RANSAC iterations to 1000.

The hyper-parameters in SUE are fined-tuned only on the Pittsburgh dataset and then fixed as $\scale=350$ and $\kNN=10$ for all datasets and experiments. An ablation over these parameters is given later in section~\ref{ablation_study}.
The SVM is trained with stochastic gradient descent with hinge loss and an L1-penalty, and a maximum of 1000 training iterations. 

\subsection{Performance comparison}
We first compare all the uncertainty estimation methods formulated in this work, both qualitatively and quantitatively, and in terms of their computational overhead.

\textbf{Area-under-the-Precision-Recall-curves:}
The AUC-PR for all the methods on all the datasets are summarized in Table~\ref{tab:aucprandtime}.
SUE outperforms other efficient methods by a clear margin, even on the Pittsburgh dataset which was used for training STUN and BTL. It is also important to note that a basic L2-distance-based uncertainty already outperforms BTL and STUN. Moreover, geometric verification outperforms all other uncertainty estimates although SUE achieves comparable performance. The precision-recall curves for the Pittsburgh dataset are shown in Fig.~\ref{fig:prcurvespitts}, and for the remainder datasets are provided in the accompanying supplementary materials.

\textbf{Computational requirements:}
We further report the time taken to compute the \textit{GV} confidence $s_q$ and the uncertainty estimates in Table~\ref{tab:aucprandtime}. 
Although \textit{GV} gives useful uncertainty estimates, 
the high computational cost of these \textit{GV} methods may be prohibitive for real-time online applications.
Our implementation of SUE is about \textit{three orders of magnitude} faster than \textit{GV} using DELF-RANSAC.

\begin{figure}
\begin{center}
\includegraphics[width=1.0\linewidth]{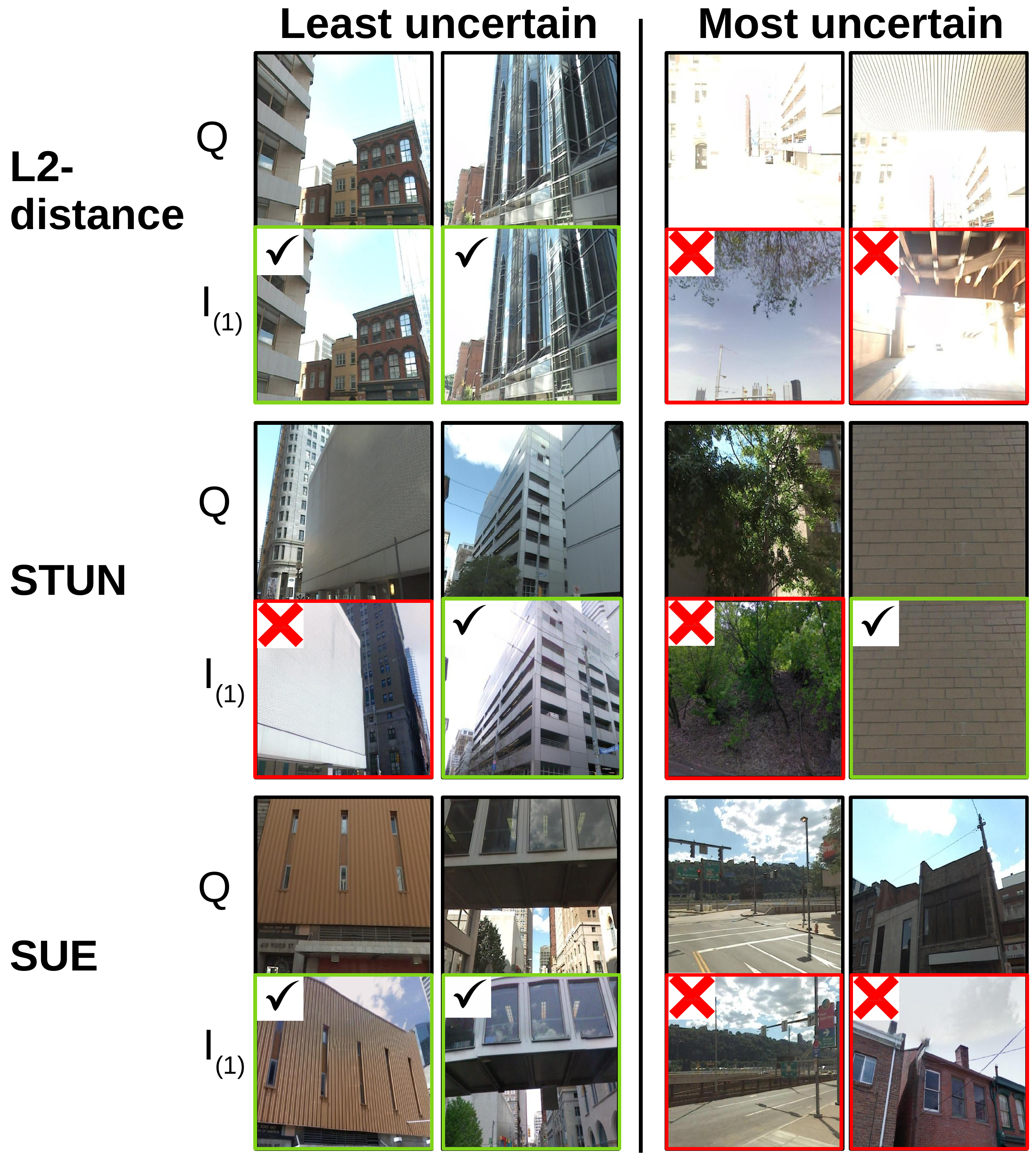}
\end{center}
   \caption{Examples of the two least and the two most uncertain query images with the corresponding nearest neighbor on the Pittsburgh dataset. The colors/symbols indicate whether the retrieved image is a correct match. }
\label{fig:exemplar_uncertainties}
\end{figure}

\textbf{Qualitative results}: 
To obtain insight into how the different methods interpret the visual content in query images and what they are sensitive to, we show 
in Fig.~\ref{fig:exemplar_uncertainties}
examples of the most and the least uncertain query images for different methods in the Pittsburgh dataset.
While all methods usually consider feature-rich and distinctive buildings as least uncertain for VPR, differences between the methods lie in the most uncertain images. Highly saturated test images are considered most uncertain by L2-distance-based uncertainty because such saturation did not exist during the reference traversals of the same scene. On the other hand, STUN considers images of trees and walls that usually contribute to perceptual aliasing as the most uncertain for VPR. SUE considers traffic squares and common building patterns as the most uncertain. Note that because SUE uses the freely available pose information in the test reference set; whether a traffic square or a building is considered uncertain is specific to this test reference set and not due to a generally-applicable visual property. 

\begin{figure}
\begin{center}
\includegraphics[width=1.0\linewidth]{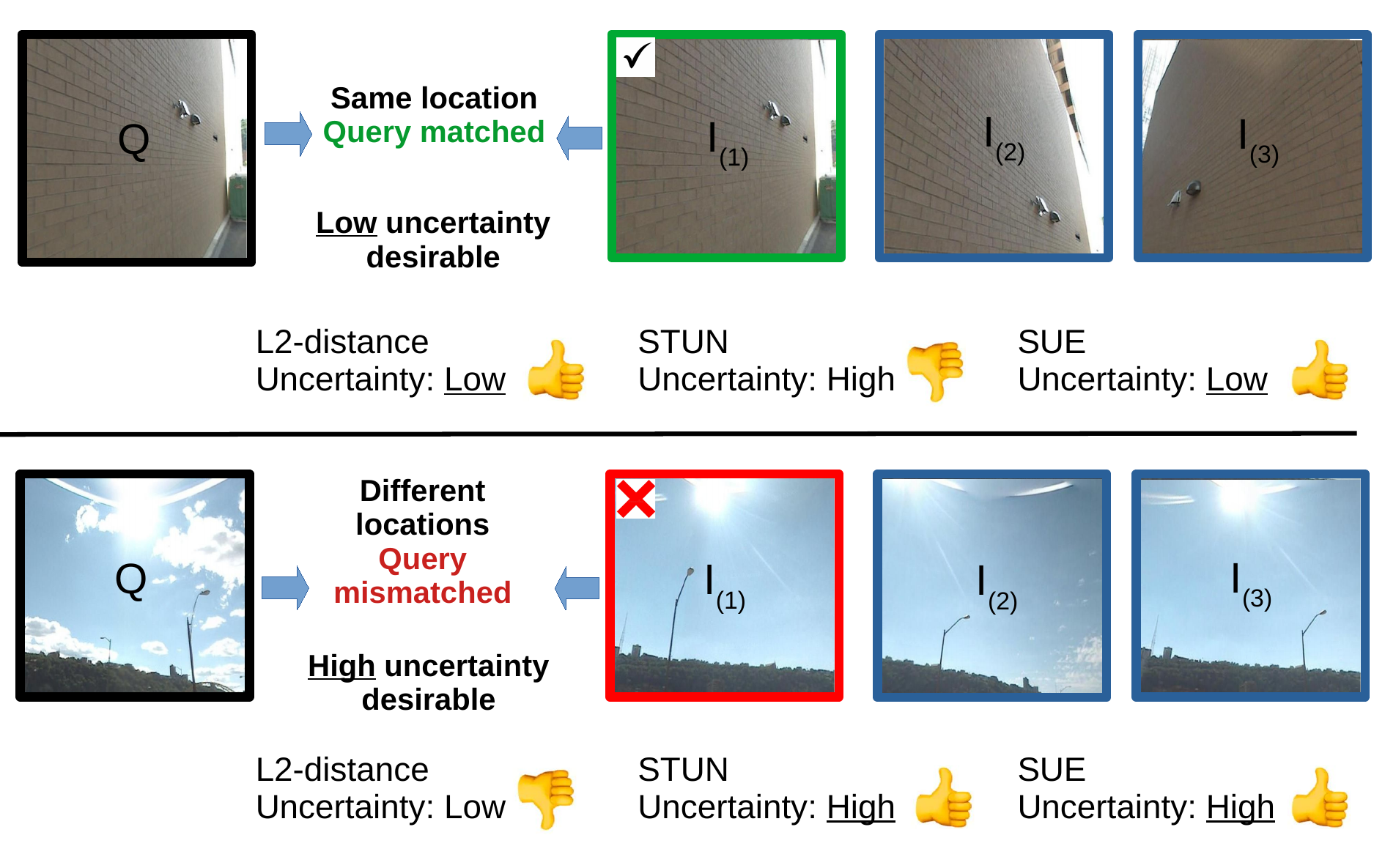}
\end{center}
   \caption{Two queries and their nearest neighbor reference images that illustrate cases where SUE outperforms other methods. 
   Ideally a method assigns high uncertainty to the mismatched query
   and low uncertainty to the correct match,
   as SUE does here.
   }
\label{fig:exemplar_uncertainty_complementarity}
\vspace{-6mm}

\end{figure}

We further show in Fig.~\ref{fig:exemplar_uncertainty_complementarity}  several images that illustrate failure cases of \textit{RUE} and \textit{DUE} in comparison to \textit{SUE}. Images of walls generally contribute to high aleatoric uncertainty (\textit{DUE}) and are closer together in the feature space in terms of L2-distance (\textit{RUE}). However, we note that the query in Fig.~\ref{fig:exemplar_uncertainty_complementarity} Top is correctly matched since only a unique wall with this pattern exists in the test reference set. SUE and L2-distance correctly give this query a low uncertainty, but STUN fails. The query image in Fig.~\ref{fig:exemplar_uncertainty_complementarity} Bottom is given low uncertainty by L2-distance than ranking with STUN and SUE. This is because images with large portions of sky contribute to aleatoric uncertainty but they are close in terms of the feature space L2-distance. This query is mismatched and identifies where L2-distance-based uncertainty fails in comparison to STUN and SUE.

\begin{figure*}
\begin{subfigure}{0.34\textwidth}
  \centering
  \includegraphics[width=0.95\linewidth,trim={0 0 0 1.22cm},clip]{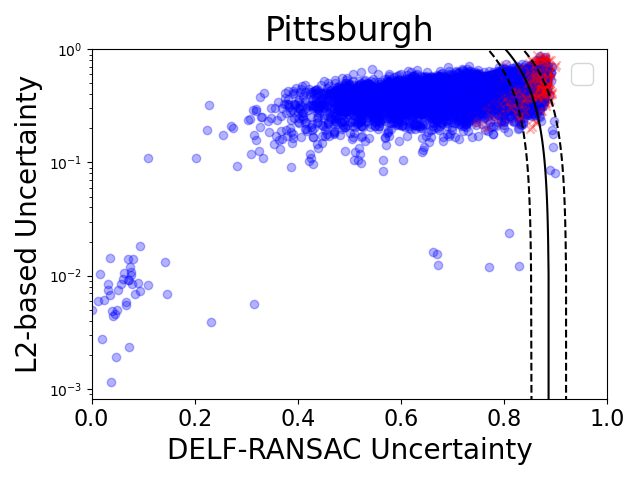}
  \label{fig:delf_l2_pitts}
\end{subfigure}%
\begin{subfigure}{0.34\textwidth}
  \centering
  \includegraphics[width=0.95\linewidth,trim={0 0 0 1.22cm},clip]{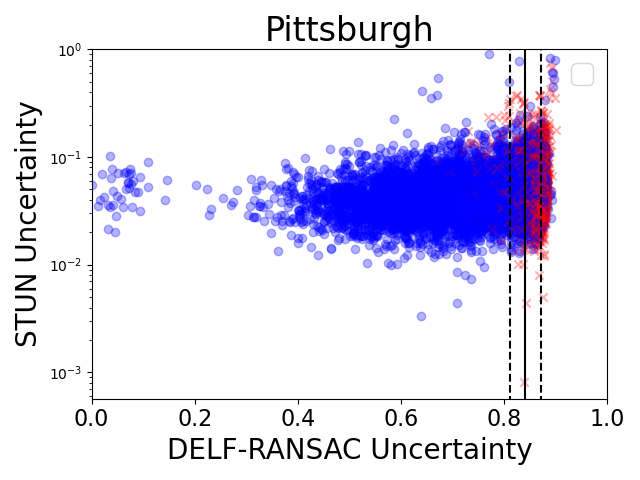}
  \label{fig:delf_stun_pitts}
\end{subfigure}
\begin{subfigure}{0.34\textwidth}
  \centering
  \includegraphics[width=0.95\linewidth,trim={0 0 0 1.22cm},clip]{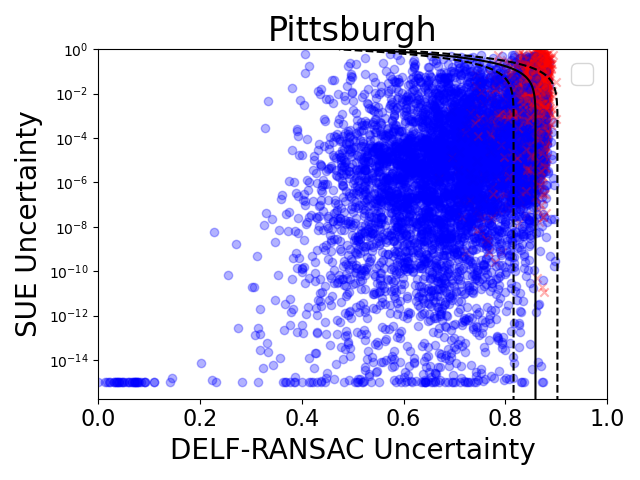}
  \label{fig:delf_ours_pitts}
\end{subfigure}
\caption{The relation between 
geometric verification uncertainty (x-axis) and the L2/STUN/SUE uncertainty (y-axis) on the Pittsburgh dataset~\cite{arandjelovic2016netvlad}.
Each point represents a query, with \textit{blue} indicating a correct match, and \textit{red} otherwise.
The linear SVM boundaries are shown as black lines, while the dashed lines are the SVM margins.
Scores have been linearly scaled to the $[0,1]$ range based on the min/max value in the training data, and for better visualization the vertical scale is in log-space, hence the SVM boundaries appear non-linear.
The class distributions in the right-most plot reveal that SUE complements geometric-verification, especially when the latter has low confidence.} 
\label{fig:uncertaintywithgv}
\vspace{-3mm}
\end{figure*}

\subsection{Complementing geometric verification}
Finally, we test if efficient uncertainty estimation can complement geometric verification,
as outlined in Sec.~\ref{sec:combiningUEwithGV}.
We show in Fig.~\ref{fig:uncertaintywithgv} the relation between the different types of uncertainties with the uncertainty from geometric verification. As we note from Table~\ref{tab:aucprandtime}, STUN outperforms BTL, and L2-distance is on average better than the PA-score, thus we only combine STUN, L2-distance, and SUE with geometric-verification for this analysis. 

\begin{figure}
\begin{center}
\includegraphics[width=1.0\linewidth]{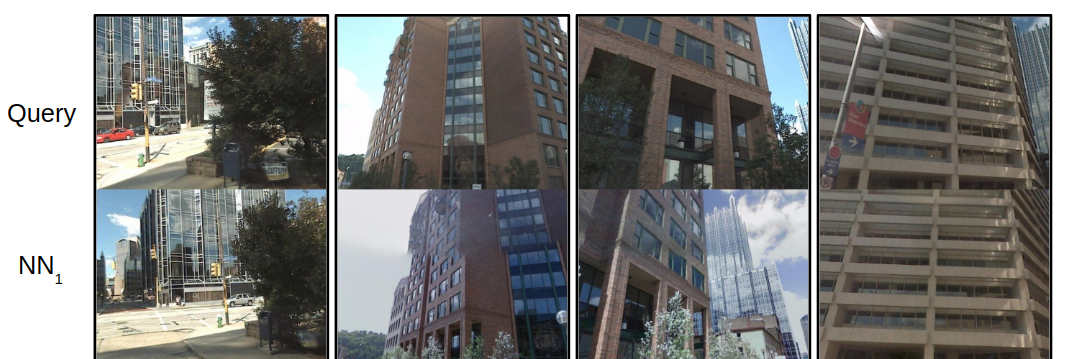}
\end{center}
   \caption{Correctly matched queries that are given high uncertainty by DELF-RANSAC and low uncertainty by SUE.}
\label{fig:complementaryqueries_to_gv}
\end{figure}

SUE provides complementary performance by giving low uncertainty $s_q$ to images that are correctly matched but which were given high \textit{GV} uncertainty. Some of these complementary queries are shown in Fig.~\ref{fig:complementaryqueries_to_gv}, where it can be seen that these queries are images that are generally difficult to match local feature descriptors, such as facades, trees, and other repetitive features within the image~\cite{knopp2010avoiding}. We also show the linear boundaries learned by SVM to classify between true-positives and false-positives. The classification accuracy of the different methods is reported in Table~\ref{tab:binaryclassacuuracy}.

\begin{table}
\begin{center}
\begin{tabular}{|l|c|c|c|c|c|}
\hline
Method &Pitts. &San. &Stlu. &Eyn. &MSLS \\ 
\hline \hline
Superpoint &85.1 &53.2 &76.4 &67.5 &36.9\\ 
DELF &86.0 &86.6 &85.3 &78.3 &\textbf{80.2}\\ 
L2-distance &75.7 &57.3 &56.2 &67.7 &36.8\\ 
STUN &74.0 &54.0 &58.0 &67.6 &37.4\\ 
\textit{SUE} &78.9 &70.7 &72.8 &77.3 &46.0\\ 
\hdashline
DELF+L2-di. &85.7 &86.1 &82.3 &77.3 &72.0\\ 
DELF+STUN &85.4 &81.6 &80.1 &75.0 &68.2\\ 
DELF+SUE &\textbf{87.1} &\textbf{89.6} &\textbf{88.7} &\textbf{82.1} &\textbf{73.4}\\ 

\hline
\end{tabular}
\end{center}
\caption{Binary classification accuracy given the uncertainty estimates of various methods, using a linear SVM trained \textit{only} on the Pittsburgh dataset. The combination \textit{DELF + SUE} generalizes better than baseline combinations, except on the MSLS dataset where although \textit{DELF+SUE} is better than the other combinations, the SVM boundaries learned from Pittsburgh are not the best.}
\label{tab:binaryclassacuuracy}
\vspace{-5mm}
\end{table}

\subsection{Ablation study}
\label{ablation_study}
\begin{figure}
\begin{subfigure}{0.23\textwidth}
  \centering
  \includegraphics[width=1.0\linewidth,trim={0 0 0 0.2cm},clip]{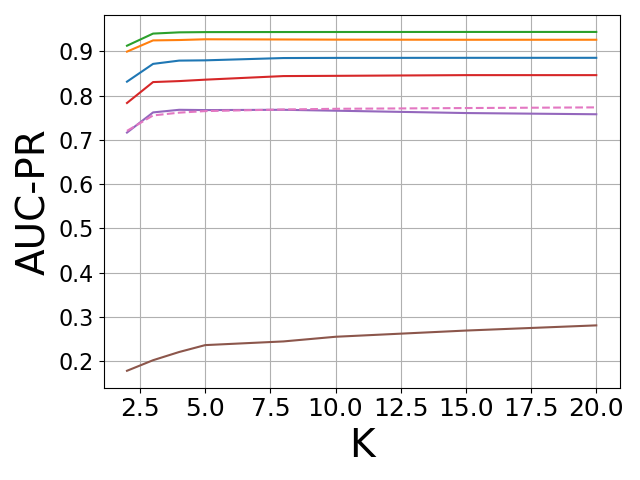}
\end{subfigure}%
\begin{subfigure}{0.23\textwidth}
  \centering
  \includegraphics[width=1.0\linewidth,trim={0 0 0 0.2cm},clip]{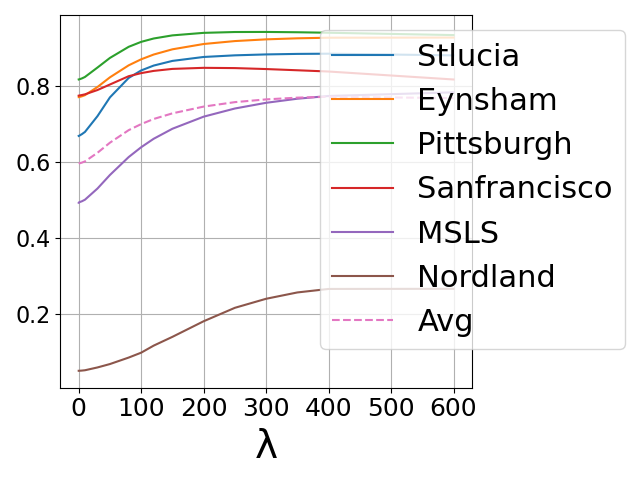}
\end{subfigure}

\caption{\camerareadychange{Effect of changing SUE's hyper-parameters $\kNN$ and $\scale$ on the AUC-PR. For each curve, the other fixed hyper-parameter is chosen as $\kNN=10$ or $\scale=350$. }}
\label{fig:ablationstudy}
\end{figure}

SUE requires two hyper-parameters, the number of nearest neighbors $\kNN$ and the decay parameter $\scale$ that controls the relative contribution of the poses of the nearest neighbors. We show the ablation over these parameters in Fig.~\ref{fig:ablationstudy} by plotting the corresponding AUC-PR values for all datasets given a set of values for each parameter. The trend remains primarily the same across all datasets. We note that the AUC-PR increases by considering more nearest neighbors but the curves mostly plateau after $\kNN=5$, since poses from low-ranked neighbors contribute less to the overall pose hypothesis. For $\scale$, we see that the range $200-400$ is generally stable and gives reliable uncertainty estimates. We also note here (with details in the appendix) that SUE generalizes to different backbones (CosPlace~\cite{berton2022rethinking}), and that exponential weighing of SUE in \camerareadychange{Equation~\eqref{eq:2}} performs better than uniform weighing (an average AUC of 0.87 vs 0.70).

\subsection{Discussion}
\label{sec:discussion}
We can now make several \textbf{recommendations} for estimating the image-matching uncertainty in VPR.
First, future works evaluating image-matching uncertainty estimation should include diverse baselines such as SUE, even if they are simple.
As our intra-category comparison revealed, even a common L2-distance-based image-matching uncertainty estimation may outperform data-driven techniques.
Second, aleatoric uncertainty from training data does not necessarily generalize to the test data,
so learning-based approaches should consider that perceptual aliasing is not just a property of the image content, but also the reference map at test time.
Referring back to the example of the sky in images being ambiguous for an outdoor map;
in an indoor map containing just one open-air patio,
such images with sky might instead be considered distinctive for their location.
Third, while \textit{GV} gives the best uncertainty estimates at the expense of high computational needs, it is still susceptible to aleatoric uncertainty within the image, 
as repetitive structures, trees, and walls may also lead to incorrect matches of local features. In VPR, \textit{GV} methods can still benefit from complementary uncertainty estimates provided by other methods, such as SUE.

We also note some potential \textbf{limitations of SUE}. 
SUE may underestimate the uncertainty if $\kNN$ is too small to retrieve aliased references from multiple locations.
Selecting $\kNN$ for maps with mixed scene depths can therefore be challenging.
Images in areas with low scene depth will already be perceptually distinct at small spatial offsets,
whereas at high scene depth even images further apart may suffer from perceptual aliasing.
A $\kNN$ that suffices for small scene depths could be too small for areas with high scene depths.
This could be mitigated by dynamically incrementing $\kNN$ till $w_{(\kNN)}$ becomes nearly zero. 
Now consider reference locations A and B which are perceptually aliased, i.e.~all their image descriptors are similar. 
If A has 1000 references and B has one,
even with $\kNN \geq 1001$, SUE will always be confident about queries from either A or B as nearly all retrieved matches are spatially close.
The high coverage of A over B thus presents an unwanted confidence bias, unless the chance of visiting A over B at test time is also $1000\times$ higher.
Nevertheless, we have shown that despite these assumptions SUE performs well on many real-world datasets. \camerareadychange{We study this more in depth in the supplementary materials, and there also present a possible solution.}



\section{Conclusions}
We have compared different approaches for estimating the image-matching uncertainty in VPR, which provided (surprising) insights into this task, e.g. existing methods that learn aleatoric uncertainty from the training dataset often do not generalize well to the reference map at test time, and the common L2-distance in the feature space can be a more reliable indicator of matching uncertainty.
We have shown that matching uncertainty in VPR is tightly related to the reference set at test time. 
Our new baseline SUE uniquely considers the spatial locations of the references,
and outperforms all but the computationally expensive geometric verification.
Its uncertainty estimates complement those of geometric verification. The choices for SUE's hyper-parameters generalize for most queries across the tested datasets.
We made recommendations for future research in this area. 

\vspace{2mm}
\camerareadychange{\noindent \textbf{Acknowledgement}. This work was supported by the 3D Urban Understanding Lab established under the TU Delft AI Initiative, and the EU Horizon 2020 programme under grant number 964505 (Epistemic AI).}

\clearpage 

{\small
\bibliographystyle{ieeenat_fullname}
\bibliography{main}
}



\section{Supplementary Material}
\camerareadychange{We provide here an ablation of SUE by changing the backbone and the weight function. A probabilistic interpretation of SUE is then presented and later used to perform density compensation for dissimilarly distributed query and reference images. We further provide the precision-recall curves for the remainder five VPR datasets. The complementarity of SUE, STUN, and L2-distance to \textit{GV} is also shown on these datasets. We also show these complementarity plots of other techniques with SUE. Then, we connect the concept of geometric burstiness~\cite{sattler2016large} with SUE. Finally, some qualitative results are shown in the form of correctly/incorrectly matched queries ranked with different types of uncertainty estimates.}

\subsection{More ablation studies of SUE}
\begin{figure*}
\begin{subfigure}{0.45\textwidth}
  \centering
  \includegraphics[width=1.0\textwidth]{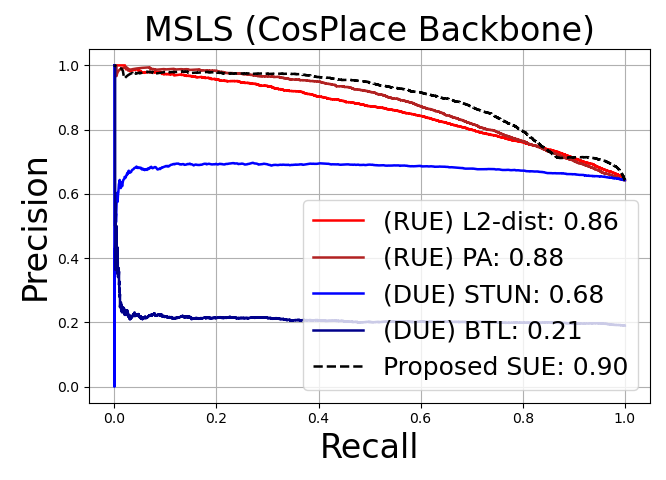}
\end{subfigure}
\hfill
\begin{subfigure}{0.45\textwidth}
  \centering
  \includegraphics[width=1.0\textwidth]{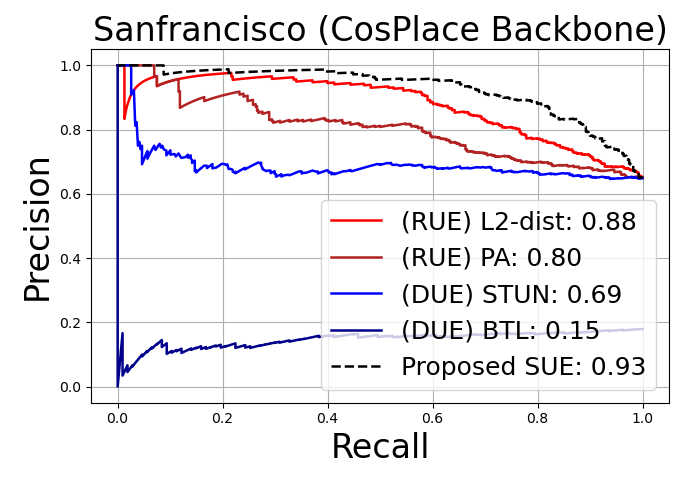}
\end{subfigure}
\caption{SUE remains SOTA by changing the backbone feature extractor to CosPlace~\cite{berton2022rethinking} with no retuning of SUE's hyper-parameters. CosPlace is also used as the backbone for L2-distance and PA-score, but it was not possible to change the backbone for BTL and STUN.}
\label{fig:more_ablation}
\end{figure*}

We perform two further experiments: changing the backbone feature extractor from STUN~\cite{cai2022stun} to CosPlace~\cite{berton2022rethinking} to show SUE's generality to other backbones in Fig.~\ref{fig:more_ablation}, and the benefit of using the exponential weighing function (in Equation~\eqref{eq:2} of the main paper) instead of the uniform weighing, as reported in Table~\ref{tab:uniformvssueweighing}.

\begin{table}[h]
\begin{center}
\begin{tabular}{|l|c|c|c|c|c||c|}
\hline
Weigh. &Pitts. &San. &Stlu. &Eyn. &MSLS & Avg\\
\hline
Uniform &0.81 &0.77 &0.67 &0.77 &0.49 &0.70\\
SUE &\textbf{0.94} &\textbf{0.84} &\textbf{0.88} &\textbf{0.93} &\textbf{0.77} &\textbf{0.87}\\
\hline
\end{tabular}
\end{center}
\caption{SUE weighs the contribution of the nearest neighbor poses based on the distance in the feature space with an exponentially decaying function. This performs better than uniform weighing of the variance of the reference poses.}
\label{tab:uniformvssueweighing}
\end{table}

\subsection{A probabilistic view of SUE}
\label{sec:probview}

\newcommand{\prob}{\mathsf{p}}
We here present a probabilistic view of SUE,
which will help formulate
a modified version
in Section~\ref{sec:correctingprior}
to account for different spatial distributions of queries and references.

Consider $M \in \{1, \cdots, N\}$ as a stochastic `match' variable that indicates which of the $N$ references is a true reference. So, $M = i$ would mean reference $i$ is the `true' match for the query.
Then $\prob(M = i)$ expresses the prior belief that any reference $i$ could be the true reference.

Assuming that some reference $i$ is the true reference, $M = i$, then the observed query feature $f_q$ can be expected to be similar to the reference feature $f_i$, with some homoscedastic Gaussian noise or variation added to all feature dimensions,
\begin{align}
\prob(f_q | M = i) &= \Gauss(f_q | f_{(i)}, \Sigma_f) \\
    &\propto e ^ {-\scale \cdot ||f_q - f_{(i)}||_2 } \\
    &\propto w_{(i)}.
\end{align}
So, the weight term of Equation~\eqref{eq:3} can be considered as the non-normalized likelihood term.
Note that the hyperparameter $\scale$ subsumes the noise parameter $\Sigma_f$.

Through Bayes' rule, we can express the posterior belief over $M$ given the query feature as
\begin{align}
\prob(M | f_q) = \frac{ \prob(f_q | M) \prob(M) }{\prob(f_q ) } = \frac{ \prob(f_q | M) \prob(M) }{ \sum_{j} \prob(f_q | M = j) \prob(M = j) }.
\label{eq:bayesrule}
\end{align}
With a uniform prior ($\prob(M) = 1/N$) that indicates equal probability for all references,
we can see that the posterior reduces to
$\prob(M | f_q) = w_{(i)} / \sum_{j} w_{(j)}$,
since the constant of the prior factors out in the numerator and denominator.

If we now assume that our VPR technique is reasonable, and that the query position should be located at the `true' reference, then we can express the expected query position, given our belief on the match of each reference, i.e., 
\begin{align}
    \mathbb{E}[p_{(M)} | f_q] &= \sum_i \left[ \prob(M = i | f_q) p_{(i)} \right] \\
    &= \mu_p
\end{align}
Here we recognise Equation~\eqref{eq:1}, assuming the uniform prior $\prob(M)$.
While we do not necessarily consider this expected pose to be representative of the true query pose (it could be an average location between distant visually-matching areas),
it does allow us to compute the expected squared pose distance of the true match to the query,
\begin{align}
    \mathbb{E}\left[||p_{(M)} - \mu_p ||_2\bigg\rvert f_q\right]
    &\approx \textrm{trace}(\Sigma_p) = s_q,
\end{align}
where $\Sigma_p$ is as defined in Equation~\eqref{eq:2} for the uniform prior $\prob(M)$.
In other words, in SUE $s_q$ estimates the expected (squared) distance between the match's pose and the query pose,
thus the smaller $s_q$ the higher the chance is that a match selected according to our posterior belief is within an acceptable distance to the true query pose.

Finally, reference $i' = \argmax_i \prob(M=i|f_q)$ with the highest posterior probability of being the correct match is selected,
which based on the likelihood term (and with uniform prior) will be $i' = 1$, i.e.~the nearest neighbor in the feature space.

Note that in the above, a uniform prior $\prob(M)$ means all references are assumed a-priori equally likely to match the query.
In case some areas in the map contain more references than other areas, this also implies a higher prior belief that the query will occur in such a denser sampled area.
This `default' prior is therefore \textit{not} a uniform \textit{spatial} prior over the mapped area, but it assumes that the local spatial density of references in the map is indicative of the probability of a query appearing in such a local region. 

\subsection{Spatial density compensation for dissimilar query/reference spatial distributions}
\label{sec:correctingprior}

As explained in SUE's potential limitations of Discussion 
Section~\ref{sec:discussion} and Appendix Section~\ref{sec:probview},
the default formulation of SUE assumes that each reference is equally probable to match a query, i.e., a uniform prior $\prob{(M)}$ is assumed.
In other words, the query and reference images/poses are expected to be  distributed similarly over the map, and the spatial density of the references in an area reflects the assumed prior probability for a query to be located in that area.

To illustrate, consider two perceptually-aliased locations A and B, where location A is represented by 100 images and location B by one image.
If a query occurs at A or B, SUE's uncertainty estimate as currently formulated in Equation~\eqref{eq:2} will be low, 
since the many references at location A will all agree on low spatial variance,
while the contribution of distant references at location B are 100$\times$ less.
This high confidence could be desired if location A is also 100$\times$ more likely to be visited at query-time than location B (i.e. the uniform $\prob(M)$ holds, so the spatial density of the references reflects a spatial prior of a query's location).
However, this prior could also be undesired if we expect queries at A and B are equally likely to occur, irrespective of the reference density.
Ultimately, what is desired depends on the application and data collection procedure.

In case the uniform prior $\prob{(M)}$ over references is undesired,
we can substitute it with a different prior
in the equations of Section~\ref{sec:probview}.
Specifically, 
in Equation~\eqref{eq:bayesrule} the likelihood terms should \textit{not} be
multiplied with a constant prior term (which cancelled out in the numerator and denominator).
Still, it may be more convenient to express the prior over references in terms of a
\textit{spatial prior for the query}.
In other words, a reference would be more probable to match if it is in a area where the query is more probable to occur, while a reference would be less probable if there are more other references in the same spatial area.
Let $\prob_q(p)$ denote the desired spatial prior for the query to be at a pose $p$,
and $\prob_r(p)$ denote the spatial density of the references at a pose $p$, then
\begin{align}
    \prob{(M = i)} \propto \frac{ \prob_q(p_{(i)}) }{ \prob_r(p_{(i)}) }.
    \label{eq:refprior}
\end{align}

We will refer to this as \textit{spatial density compensation}.
In practice, we can thus compensate SUE for a desired spatial prior by multiplying the reference weight $w_{(i)}$ with a term (proportional to) the desired prior $\prob{(M)}$.
Note from Equation~\eqref{eq:refprior} that if the spatial distributions of queries and references are assumed equal, we again obtain that $\prob{(M)}$ is uniform,
as is the case for the default SUE formulation.


\begin{figure}
    \centering
    \begin{subfigure}[b]{0.23\textwidth}
        \centering
        \includegraphics[width=\textwidth,trim={0 0 0 1.5cm},clip]{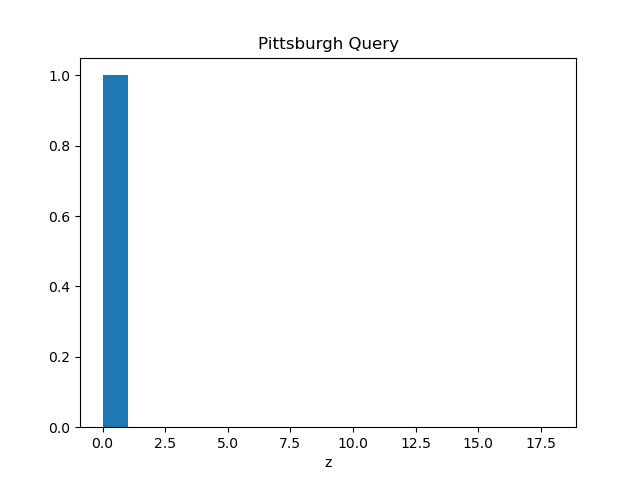}
        \caption[]%
        {{\small Pittsburgh query}}    
    \end{subfigure}
    \hfill
    \begin{subfigure}[b]{0.23\textwidth}  
        \centering 
        \includegraphics[width=\textwidth,trim={0 0 0 1.5cm},clip]{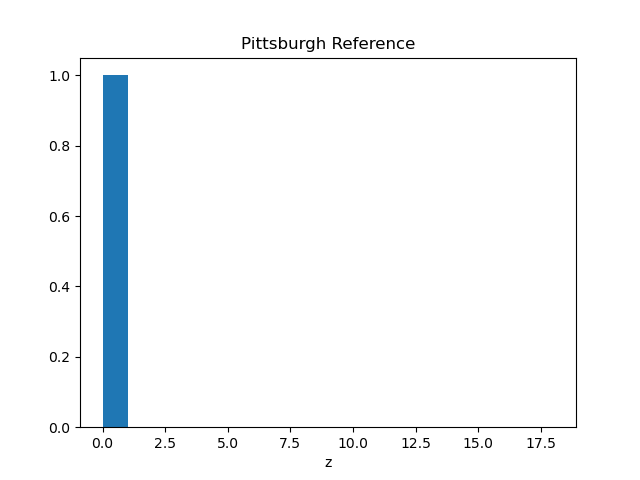}
        \caption[]%
        {{\small Pittsburgh reference}}    
    \end{subfigure}
    \hfill    
    \begin{subfigure}[b]{0.23\textwidth}   
        \centering 
        \includegraphics[width=\textwidth,trim={0 0 0 1.5cm},clip]{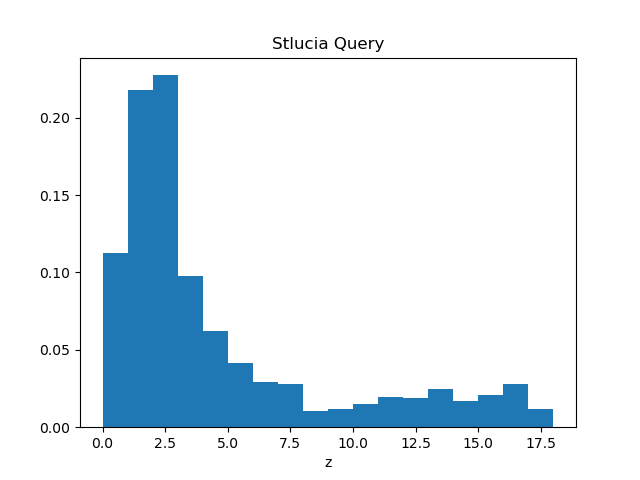}
        \caption[]%
        {{\small Stlucia query}}    
    \end{subfigure}
    \hfill
    \begin{subfigure}[b]{0.23\textwidth}   
        \centering 
        \includegraphics[width=\textwidth,trim={0 0 0 1.5cm},clip]{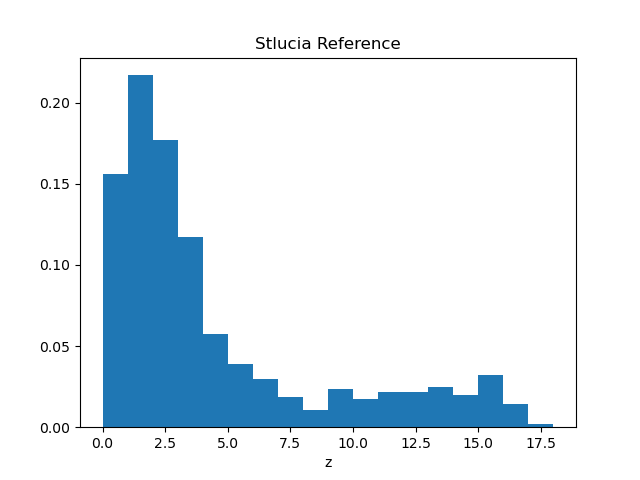}
        \caption[]%
        {{\small Stlucia reference}}    
    \end{subfigure}
    \caption[]
    {\camerareadychange{\small The density of queries and references is depicted using the distance ($z$) of each query/ref to its nearest neighbour ($k=1$) in the pose space. Queries and references in Pittsburgh dataset are highly dense and hence uniformly spatially distributed. The queries and references are non-uniformly (albeit similarly) spatially distributed in the sparser Stlucia dataset.} }
    \label{fig:queryrefdist}
\end{figure}

\subsection{Validating spatial density compensation}
In this section, we test the spatial density compensation concept of adjusting SUE as explained in Section~\ref{sec:correctingprior}.

\paragraph{Applying a uniform spatial prior for the query}
Let's assume the spatial density of query poses is uniform,
so all query poses within the map are equally likely,
in which case term $\prob_q(p)$ becomes a constant (and thus will cancel out when normalizing the weights).

The spatial density of the references $\prob_r(p)$ can be estimated from the finite samples of poses in the reference set.
We can for instance model the spatial density of references by simply taking the distance $z_{(i)}$ of the reference $i$ to its $k$-th nearest neighbor in the \textit{pose space}, such that the area $z_{(i)}^2$ is inversely proportional to the local density of the reference $i$, i.e.,
~$\prob_r(p_{(i)}) \propto  1/z_{(i)}^2$.
Hyperparameter $k$ regularizes the smoothness of the estimated reference pose density.

We can now see that
$\prob(M = i) \propto z_{(i)}^2$,
thus the density compensated SUE for this uniform spatial prior for query poses is obtained by re-weighing Equation~\eqref{eq:3} with $z_{(i)}^2$, i.e., 
\begin{align}
    w_{(i)} &=  {e ^ {-\scale \cdot d_{(i)}}} \cdot z_{(i)}^2.
\label{eq:denscompSUE}
\end{align}


\paragraph{Do common datasets have a uniform query distribution?}
We used the above formulation of spatial density
to study the properties in the used VPR datasets.
First, we find that most of our datasets \textit{do} have a  mostly uniform spatial distribution for both queries and references, except the Stlucia dataset.
Fig.~\ref{fig:queryrefdist} illustrates the distribution of distances to the $k=1$ nearest neighbors for the Pittsburgh and Stlucia datasets.
Second, we can conclude that the assumption that references and queries have a similar spatial distribution \textit{does hold} in common VPR dataset,
hence SUE's default formulation with uniform reference prior is reasonable.

To properly validate the density compensation concept of Section~\ref{sec:correctingprior}, we also create a modified version of the Stlucia data such that queries and reference actually do have a \textit{different} spatial distribution.
We greadily subsample the Stlucia queries such that the spatial density of the resampled queries is uniform.

\paragraph{Does assuming a uniform query distribution help?}

Finally, we test
the density compensated SUE of Equation~\eqref{eq:denscompSUE}
on the VPR datasets
for different choices of $k$,
see Table~\ref{tab:compensation}.


Since queries and references of datasets other than Stlucia are already uniformly distributed spatially, the table confirms that density compensation does not lead to any major effect on SUE's performance.
We also see that for the (unmodified) Stlucia dataset, density compensation actually \textit{hurts} performance because the queries and references 
are in fact \textit{non}-uniformly and similarly distributed. 
The default uniform prior assumption of SUE is therefore better suited for Stlucia.

However, 
if we test density compensated SUE on the modified Stlucia dataset
where queries are in fact uniformly spatially distributed
while the references are not,
then we do observe a benefit over the default SUE
as shown in Table~\ref{tab:compensationforuniformstlucia}.
In this case, the spatial prior of density compensated SUE does hold,
where as the default SUE assumption
that queries and references are similarly distributed does not.

\begin{table}
    \centering
    \begin{tabular}{|c|c|c|c|c|c|}
\hline
\textbf{Compensation} &Pitts.	&San.	&Stlu.	&Eyns. &MSLS	\\
\hline
none &0.94	&0.84	&0.89	&0.93 &0.76	\\
$k=1$	    &0.94	&0.84	&0.82	&0.93 &0.76	\\
$k=3$	    &0.94	&0.84	&0.84	&0.93 &0.77	\\
$k=10$	&0.94	&0.81	&0.85	&0.92 &0.77	\\
\hline    
    \end{tabular}
    \vspace{-3mm}
    \caption{\camerareadychange{SUE's AUC-PR with reference density compensation.}}
    \label{tab:compensation}
\end{table}

\begin{table}
    \centering
    \begin{tabular}{|c||c|c|c|c|c|c|}
\hline
\textbf{\textbf{$z$}} &none	&k=1	&k=3	&k=5	&k=8 &k=10 \\
\hline
$8-9$	   &0.92	&\textbf{0.96}	&0.96	&0.96	&0.94	&0.94\\

$10-11$	   &0.68	&\textbf{0.76}	&0.73	&0.7	&0.71	&0.69\\
\hline    
    \end{tabular}
    \vspace{-3mm}
    \caption{\camerareadychange{SUE's AUC-PR with reference density compensation using different values of $k$ on the Stlucia dataset when the queries are resampled to have a close to uniform spatial density (e.g., $z=8-9$). Reference density compensation helps SUE when queries are spatially uniformly distributed and references are non-uniformly distributed. Best across the columns is in Bold.}}
    \label{tab:compensationforuniformstlucia}
\end{table}

In conclusion, 
whether spatial density compensation is needed depends on the specifc spatial distributions of the references and queries in a dataset. 
For the studied VPR benchmark datasets that represent densely collected queries and references, the default assumption of SUE 
that their spatial distributions are similar
holds.
Still, in applications where we can expect that queries and references are distributed differently, then additional density compensation can be helpful.
The formulation of spatial density compensation can be motivated from a probabilistic view on SUE.
Future work can investigate better estimates for query and reference density for non-uniformly distributed data to further improve SUE.

\subsection{Precision-Recall curves}
\begin{figure*}
\begin{subfigure}{0.45\textwidth}
  \includegraphics[width=\textwidth]{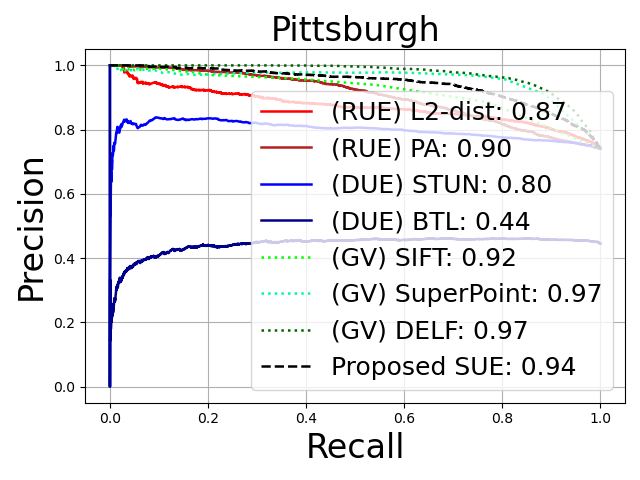}
\end{subfigure}%
\hfill
\begin{subfigure}{0.45\textwidth}
  \includegraphics[width=\textwidth]{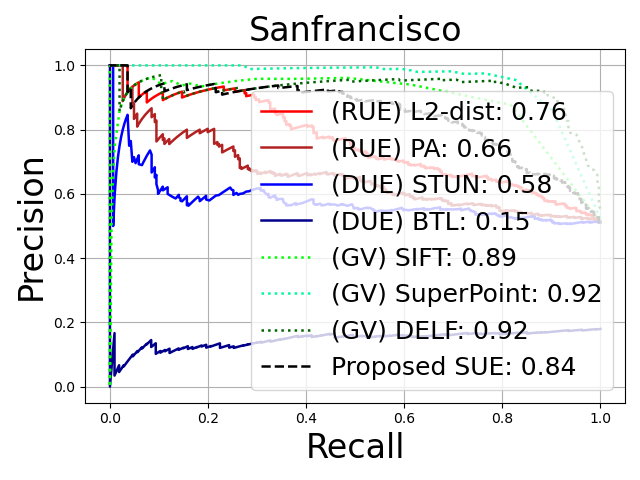}
\end{subfigure}%
\hfill
\begin{subfigure}{0.45\textwidth}
  \includegraphics[width=\textwidth]{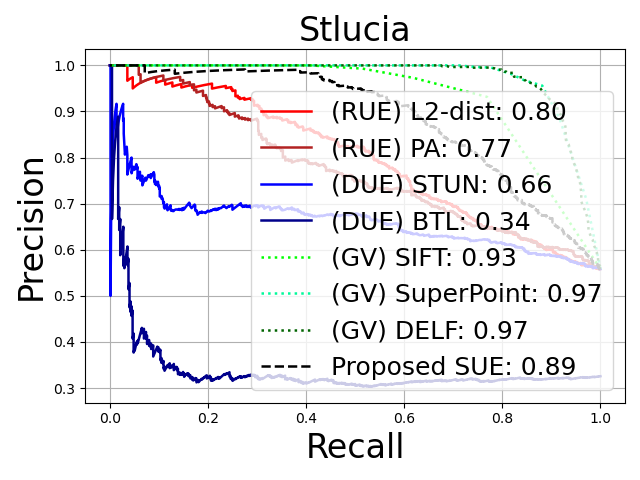}
\end{subfigure}%
\hfill
\begin{subfigure}{0.45\textwidth}
  \includegraphics[width=\textwidth]{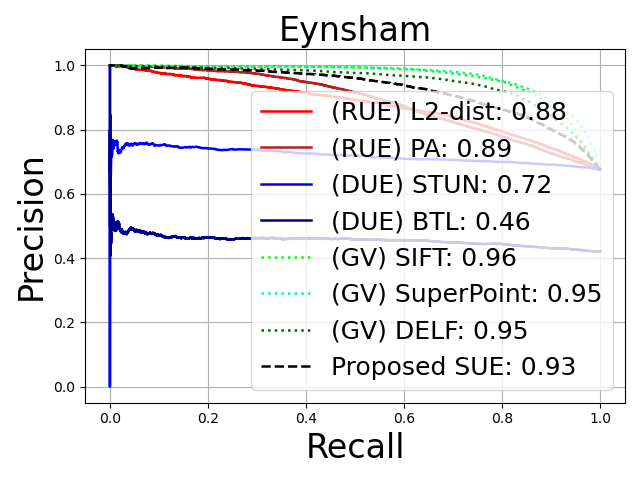}
\end{subfigure}
\hfill
\begin{subfigure}{0.45\textwidth}
  \includegraphics[width=\textwidth]{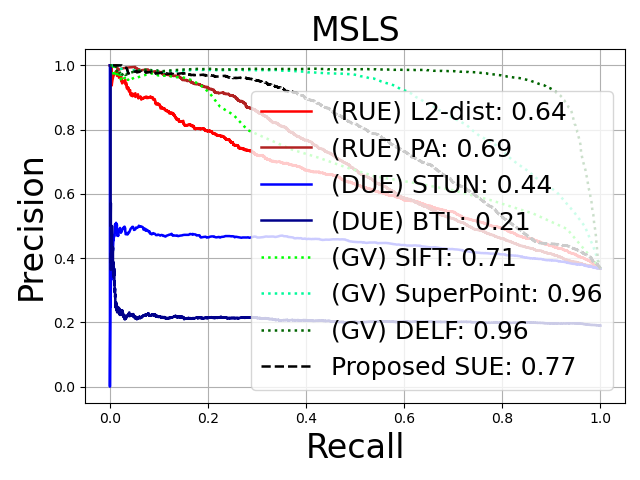}
\end{subfigure}
\hfill
\begin{subfigure}{0.45\textwidth}
  \includegraphics[width=\textwidth]{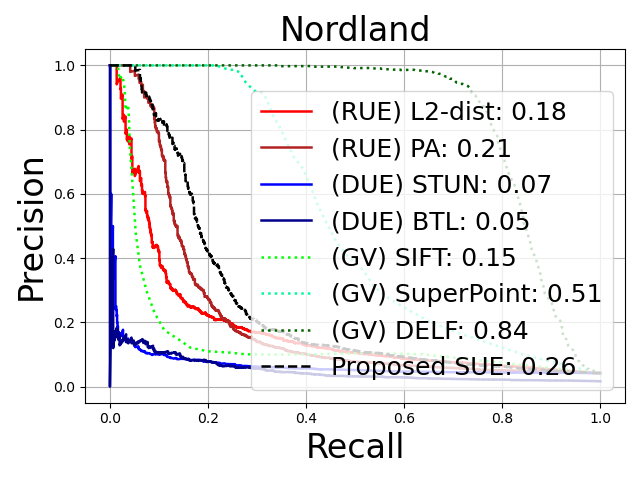}
\end{subfigure}%
\caption{\camerareadychange{The precision-recall curves on the six datasets using SUE and other baselines. 
SUE outperforms the existing methods within the efficient category on all datasets. Note how an L2-based retrieval uncertainty outperforms the data-driven aleatoric uncertainty estimated in BTL and STUN.}}
\label{fig:prcurves}
\end{figure*}

In addition to the Precision-Recall curves of the Pittsburgh dataset in Fig.~\ref{fig:prcurvespitts}, the PR-curves for the remainder five datasets are shown in Fig.~\ref{fig:prcurves}. SUE outperforms the methods in the \textit{RUE} and \textit{DUE} categories on all datasets. \textit{GV} remains the overall state-of-the-art, albeit at a two to three orders of magnitude higher computational cost. 

\subsection{Complementing geometric verification} 
\begin{figure*}
\begin{subfigure}{0.33\textwidth}
  \centering
  \includegraphics[width=0.95\linewidth]{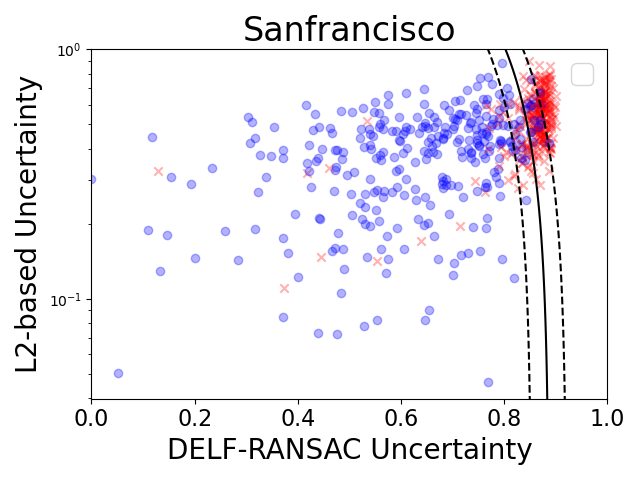}
\end{subfigure}%
\begin{subfigure}{0.33\textwidth}
  \centering
  \includegraphics[width=0.95\linewidth]{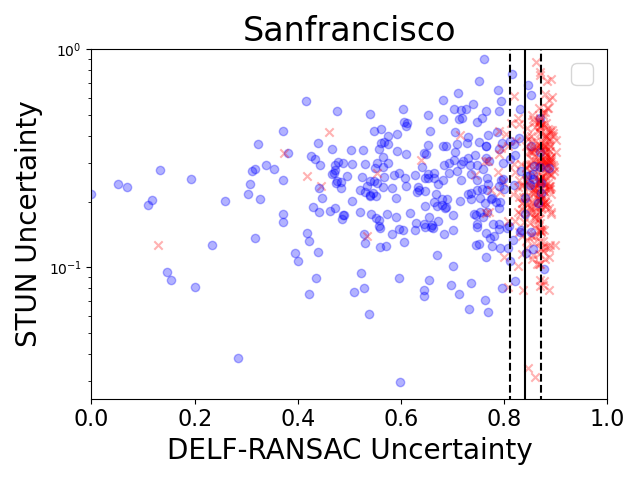}
\end{subfigure}
\begin{subfigure}{0.33\textwidth}
  \centering
  \includegraphics[width=0.95\linewidth]{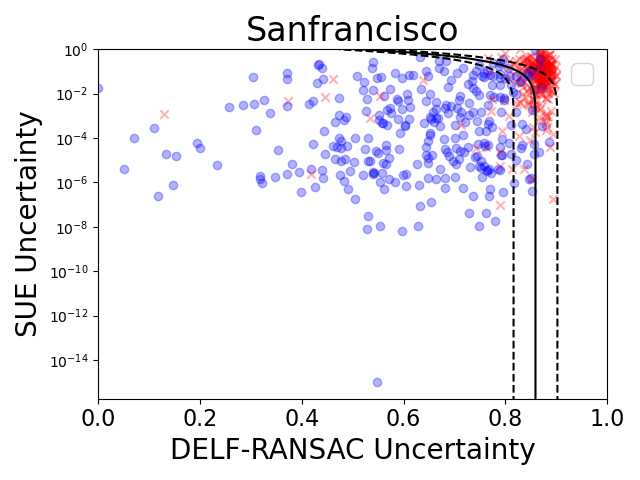}
\end{subfigure}
\begin{subfigure}{0.33\textwidth}
  \centering
  \includegraphics[width=0.95\linewidth]{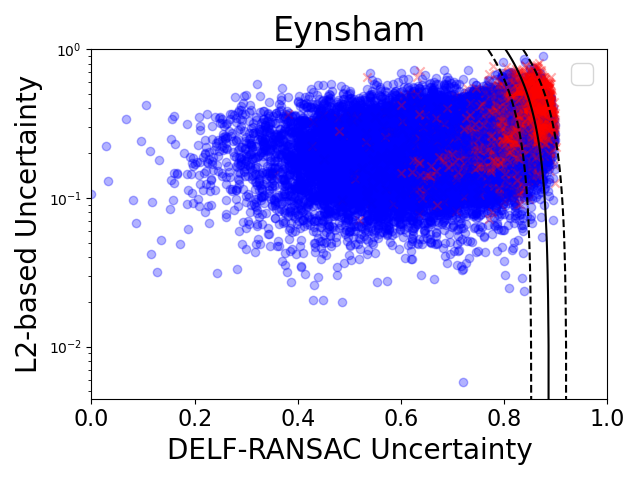}
\end{subfigure}%
\begin{subfigure}{0.33\textwidth}
  \centering
  \includegraphics[width=0.95\linewidth]{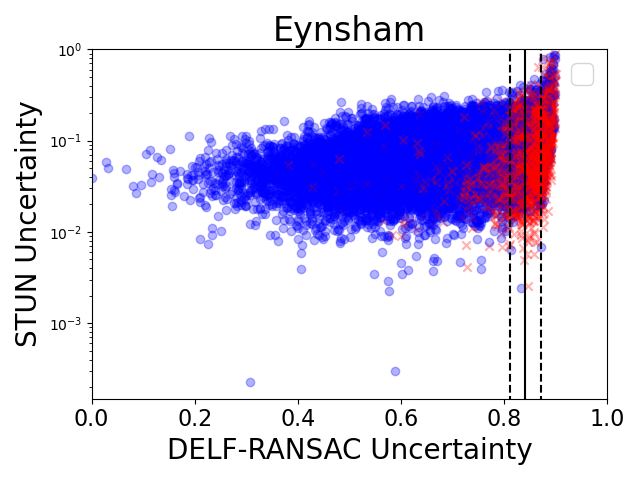}
\end{subfigure}
\begin{subfigure}{0.33\textwidth}
  \centering
  \includegraphics[width=0.95\linewidth]{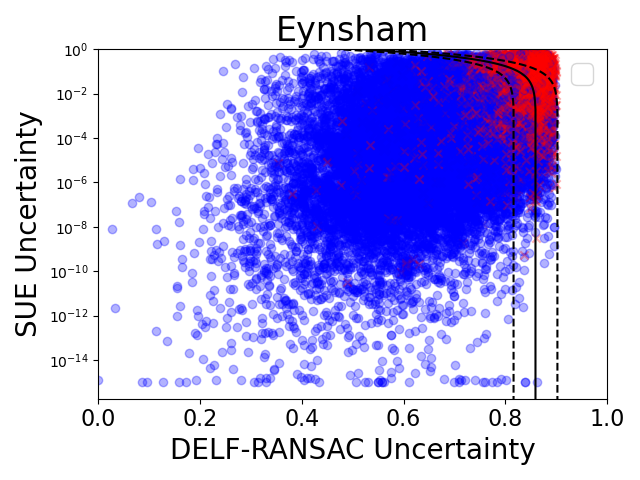}
\end{subfigure}
\begin{subfigure}{0.33\textwidth}
  \centering
  \includegraphics[width=0.95\linewidth]{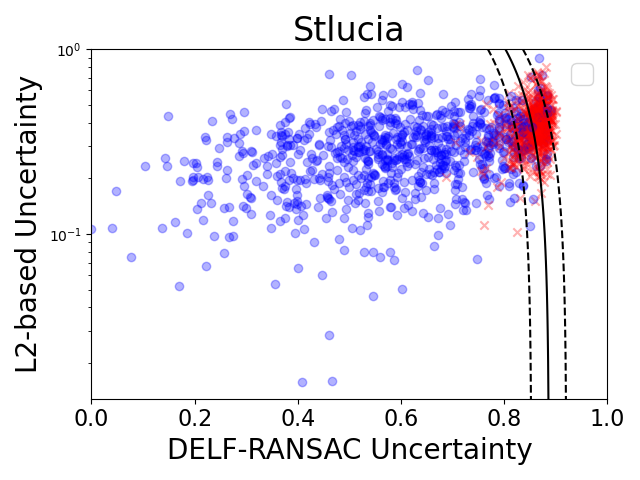}
\end{subfigure}%
\begin{subfigure}{0.33\textwidth}
  \centering
  \includegraphics[width=0.95\linewidth]{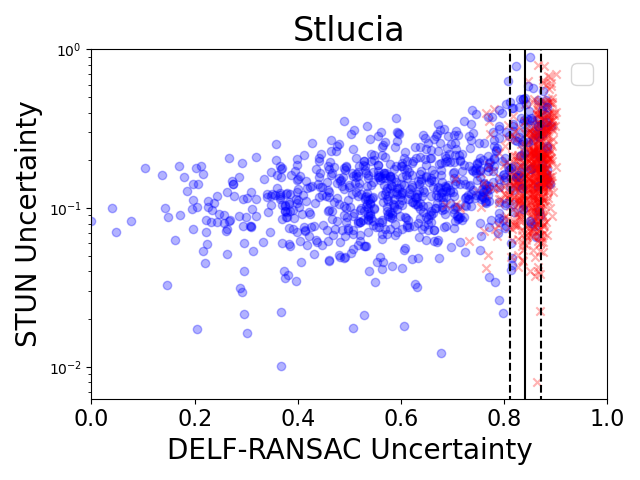}
\end{subfigure}
\begin{subfigure}{0.33\textwidth}
  \centering
  \includegraphics[width=0.95\linewidth]{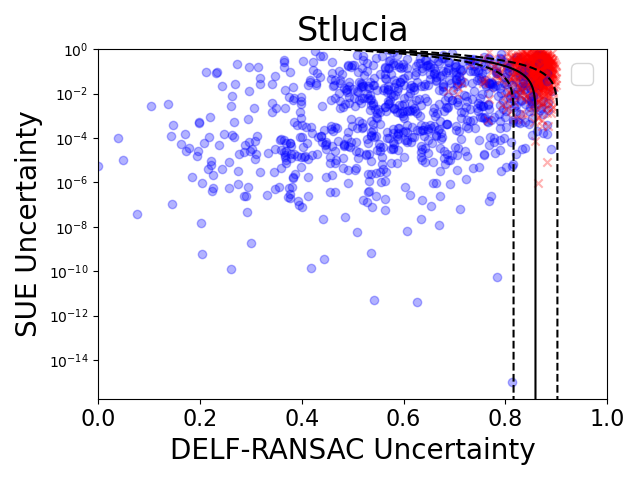}
\end{subfigure}
\begin{subfigure}{0.33\textwidth}
  \centering
  \includegraphics[width=0.95\linewidth]{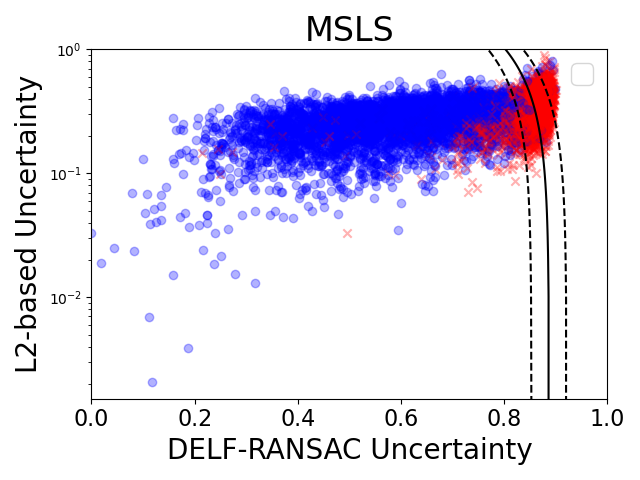}
\end{subfigure}%
\begin{subfigure}{0.33\textwidth}
  \centering
  \includegraphics[width=0.95\linewidth]{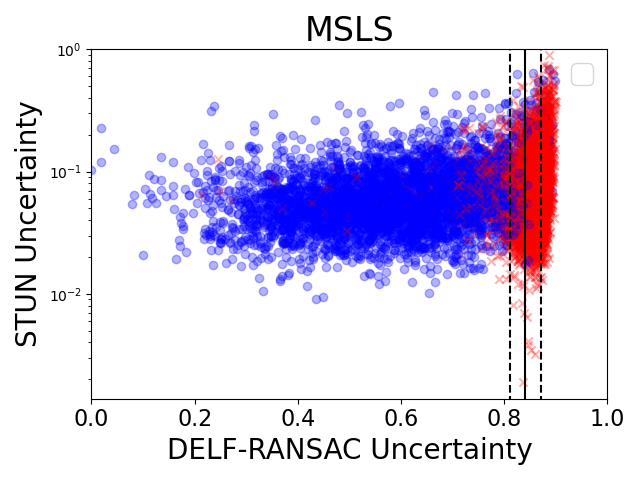}
\end{subfigure}
\begin{subfigure}{0.33\textwidth}
  \centering
  \includegraphics[width=0.95\linewidth]{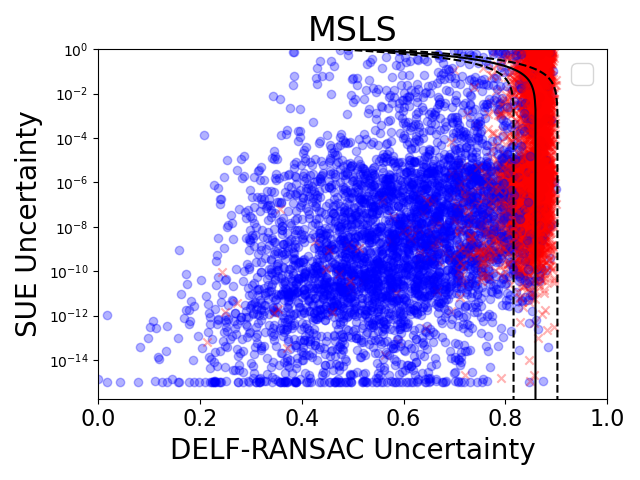}
\end{subfigure}
\caption{The relation of L2-based uncertainty, STUN, and SUE with geometric verification uncertainty. The SVM boundaries are learned on the Pittsburgh dataset only. Each point represents a query, and the color indicates whether it is a true-positive (Blue) or a false-positive (Red). The linear SVM boundaries are shown as black lines, while the dashed lines are the SVM margins. The combination of SUE with geometric-verification leads to more correctly matched queries in the bottom right (where SUE is certain but \textit{GV} is uncertain) of the plots identifying complementarity. For better visualization, the vertical scale is in log-space, due to which the SVM boundaries appear non-linear to the reader but are linear.}
\label{fig:uncertaintywithgv_FULL}
\end{figure*}

We further show in Fig.~\ref{fig:uncertaintywithgv_FULL} the generalization of SVM trained on the Pittsburgh dataset to other datasets. For all these datasets, the relation of our SUE uncertainty with DELF-RANSAC leads to complementarity with queries in the bottom-left of the plot that can be linearly separated. 


\subsection{\camerareadychange{\textbf{SUE combined with other uncertainty estimates}}}
\begin{figure*}
\centering
\begin{subfigure}{0.45\textwidth}
  \includegraphics[width=\textwidth,trim={0 0 0 1.2cm},clip]{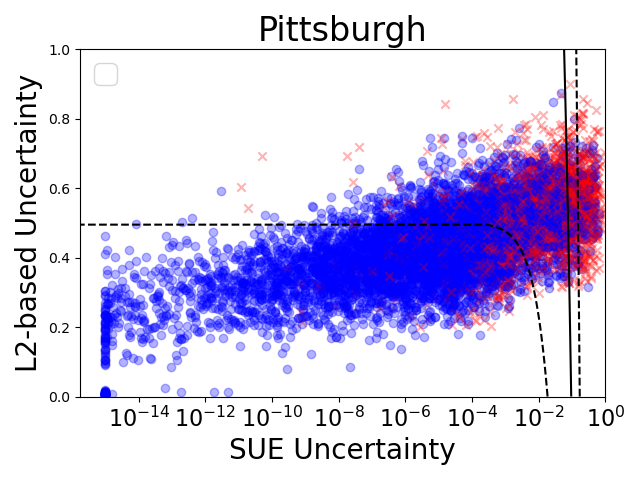}
\end{subfigure}%
\hfill
\begin{subfigure}{0.45\textwidth}
  \includegraphics[width=\textwidth,trim={0 0 0 1.2cm},clip]{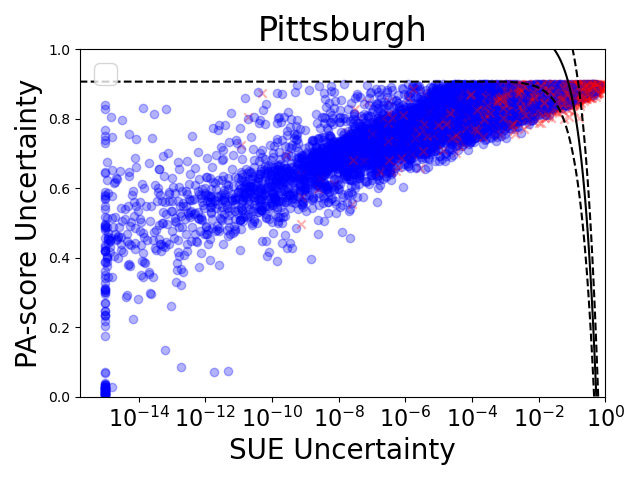}
\end{subfigure}
\hfill
\begin{subfigure}{0.45\textwidth}
  \includegraphics[width=\textwidth,trim={0 0 0 1.2cm},clip]{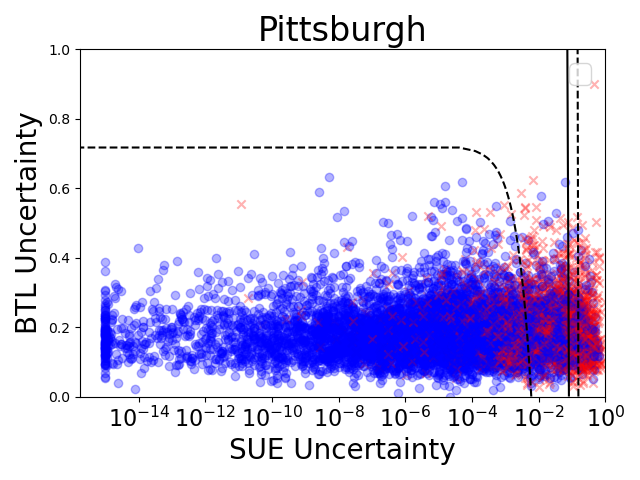}
\end{subfigure}
\hfill
\begin{subfigure}{0.45\textwidth}
  \includegraphics[width=\textwidth,trim={0 0 0 1.2cm},clip]{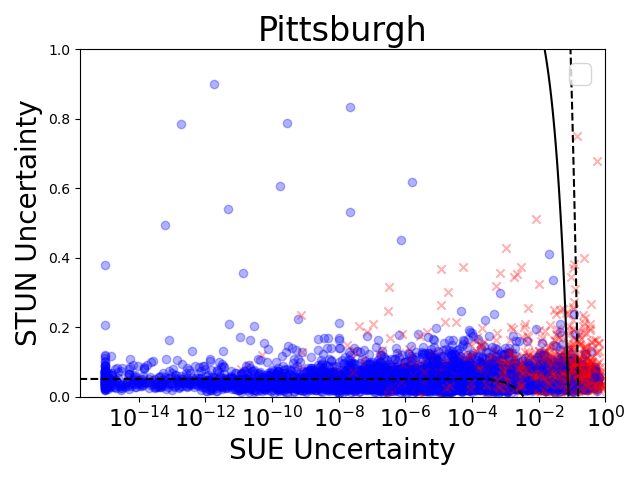}
\end{subfigure}
\hfill
\caption{\camerareadychange{The relation of L2-based, PA-score, BTL, and STUN uncertainties with SUE uncertainty. Each point represents a query, and the color indicates whether it is a true-positive (Blue) or a false-positive (Red). The linear SVM boundaries are shown as black lines, while the dashed lines are the SVM margins. As indicated by the near-vertical decision boundaries, most of the queries that can be classified as true- or false-positives by other methods can also be classified by SUE, and we do not see much complementarity.}}
\label{fig:otheruncertaintywithSUE}
\end{figure*}

For completeness, we show the combination of other uncertainty estimation methods with SUE in Fig.~\ref{fig:otheruncertaintywithSUE}. Most of the queries that can be classified as true- or false-positives by other methods can already be classified using only SUE. We hypothesize that this is because of SUE's similarity to BTL and STUN which also estimate the aleatoric uncertainty, and since SUE already uses the L2-distance and nearest neighbours in its uncertainty estimate.

\subsection{Relating SUE to geometric burstiness}
\textbf{Relation:} Features that appear in similar configurations across multiple unrelated reference images are referred to as geometric burstiness (GB)~\cite{sattler2016large}. Ideally, such features should not be considered for estimating the image matching confidence using geometric verification (GV). Whether images are related or unrelated is determined using their pose information, i.e., different images that are physically close to each other could be looking at the same place. While the use of pose information of the Top-$K$ retrieved reference images is common between SUE and GB, the latter is evaluated for image re-ranking and the former for VPR. \textit{GB} is implemented on top of \textit{GV} and is more computationally expensive than GV, concretely by an order of $K$, but gives better uncertainty estimates. For completeness, we implement a version of \textit{GB} inspired by~\cite{sattler2016large} and compare it to SUE. The implementation details are as follows. \\

\noindent \textbf{Our implementation of GB:} We use SIFT features, and perform feature matching in a RANSAC loop between a query and its Top-$K$ retrieved nearest neighbors. Local feature matches [$q_i, r^k_j$] that satisfy a geometric transform (homographic) are considered inliers, where $q_i$ is the $i$th query feature and $r^k_j$ is the $j$th feature in $k$th nearest neighbour. A query feature $q_i$ contributes to geometric burstiness if it forms part of the inlier set for multiple (say $T$) retrieved images, and in the most naive case, such [$q_i, r^k_j$] should be discarded from the inlier count. But similar to~\cite{sattler2016large}, we down-weight their contribution by $T$ instead of completely discarding such inliers.

However, Sattler~\textit{et al.}~\cite{sattler2016large} further studied that the top retrieved images could come from the same place, and hence query features could \textit{legally} form part of the inlier set for multiple retrieved images. To classify whether a set of reference images represents the same place or not, we use the definition of place from~\cite{berton2022deep} where images that are within 25 meters of each other are considered as the same place. Thus, only inliers from reference images of different (more than 25 meters apart) places are classified as geometric bursts. We use $K=20$ and for feature matching the same hyperparameters are used as that of SIFT-RANSAC. \\

\noindent \textbf{Results:}
We report in Table~\ref{tab:GBresults} that adding \textit{GB} on top of SIFT-RANSAC leads to better performance than just using SIFT-RANSAC. Overall, among all uncertainty estimation methods, DELF-RANSAC still performs the best. \textit{GB} is the most computationally expensive among all the uncertainty estimation methods. Note that \textit{GB} could also be added on top of Superpoint-RANSAC and DELF-RANSAC albeit at an even higher computational cost. 

\begin{figure}
\begin{center}
\includegraphics[width=1.0\linewidth]{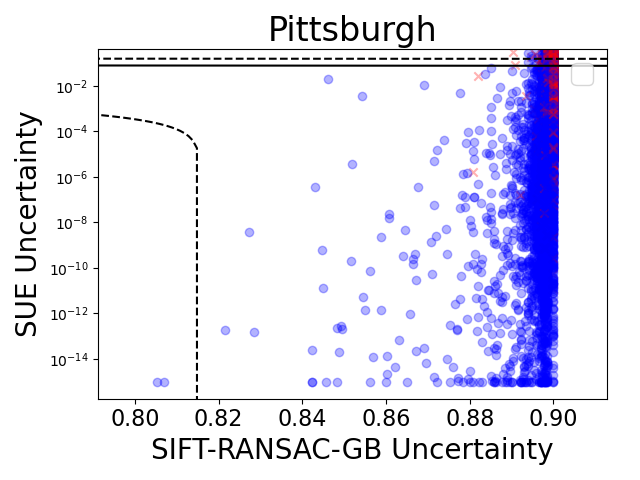}
\end{center}
   \caption{SUE remains complementary to \textit{GB} since many true-positives can be separated from false-positives using SUE uncertainty and not using \textit{GB}. See other such plots in this paper for details on the employed info-graphics.}
\label{fig:sue_with_siftransacgb}
\end{figure}

We further test if SUE remains complementary to GB, given that both methods use reference poses. Fig.~\ref{fig:sue_with_siftransacgb} shows that the uncertainty estimates from SUE can also complement \textit{GB}. In conclusion, the several orders of magnitude higher computational needs of \textit{GB} compared to SUE, and their mutual complementarity suggest that SUE is a useful baseline for uncertainty estimation in VPR.

\begin{table}
    \centering
    \begin{tabular}{|c||c|c|c|c|c|c|}
\hline
 & $\uparrow$ Pitts.	& $\uparrow$ Nord.	& $\uparrow$ MSLS	& $\downarrow$ Time	\\
\hline
L2-dist	   &0.87	&0.18	&0.64	&\textbf{0.05}\\
STUN	   &0.79	&0.05	&0.44	&0.10\\
SUE	   &0.94	&0.26	&0.77	&1.08\\
SIFT   &0.92	&0.15	&0.70	&129\\
DELF	   &\textbf{0.97}	&\textbf{0.84}	&\textbf{0.95}	&1587\\
GB (SIFT)	   &0.92	&0.31	&0.87	&2709\\
\hline    
    \end{tabular}
    \caption{\camerareadychange{AUR-PR and computation time (msecs) comparison of the methods discussed in the main paper with geometric burstiness~\cite{sattler2016large}. Best across the columns is in Bold. Implementing \textit{GB} on top of SIFT-RANSAC leads to better performance but at several orders of magnitude higher computational cost.}}
    \label{tab:GBresults}
\end{table}

\subsection{Qualitative results} 
\begin{figure*}
\begin{center}
\includegraphics[width=1.0\linewidth]{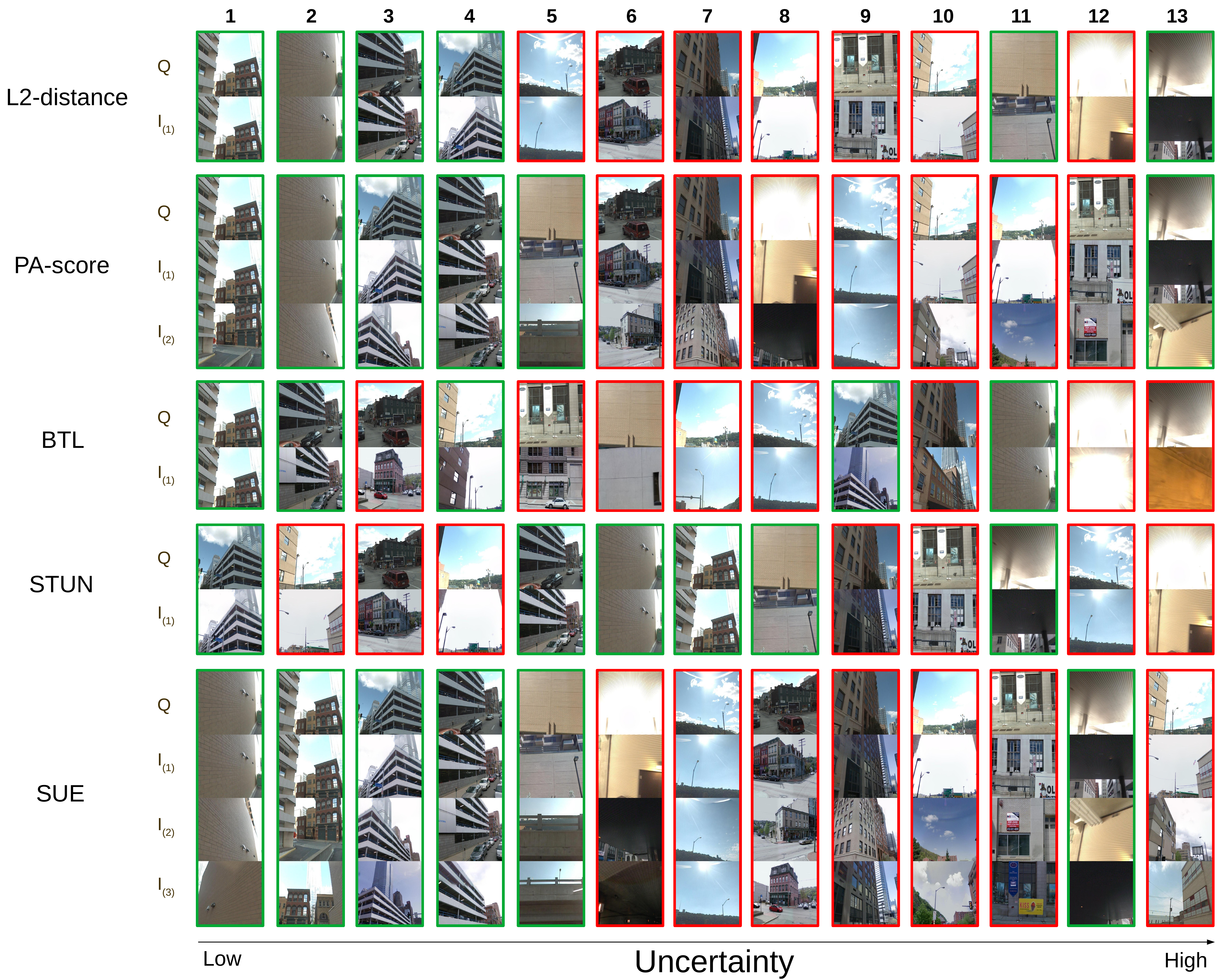}
\end{center}
   \caption{Exemplar matched/mismatched queries are ranked with different types of estimated uncertainties in the Pittsburgh dataset. Note that the set of chosen queries is the same for all types of uncertainty estimation methods. $I_{(n)}$ denotes the nearest neighbor where the subscript $n$ denotes its rank.
   The number of nearest neighbors shown relates to the corresponding number needed by each method (e.g. PA-score requires two nearest neighbors).
   The retrieved nearest neighbors for BTL are different than other methods due to the different feature encoder. A good uncertainty estimation method when used for ordering would rank correct matches to the left and incorrect matches to the right of the reader. The query image in column 12 of SUE depicts the \underline{failure case of SUE}, where the perceptually aliased nearest neighbors are geographically far-apart leading to high uncertainty but the best match is still the correct match.}
\label{fig:exemplar_queries}
\end{figure*}

We show examples of queries with their corresponding nearest neighbors ranked with the uncertainties computed by the different types of uncertainty estimation methods in Fig.~\ref{fig:exemplar_queries}. We keep the set of randomly chosen queries the same for all the methods. These examples further indicate what each method is sensitive to for uncertainty estimation. 

\end{document}